\documentclass[11pt]{article}

\usepackage{acl}

\usepackage{times}
\usepackage{latexsym}
\usepackage{booktabs}
\usepackage{listings}
\usepackage{subfigure}
\usepackage{authblk}
\usepackage{amsmath}
\usepackage{amssymb}
\usepackage{subcaption}
\usepackage[table]{xcolor}
\usepackage{colortbl}
\usepackage[most]{tcolorbox}
\usepackage[T1]{fontenc}
\usepackage[utf8]{inputenc}

\usepackage{microtype}
\usepackage{xspace}
\usepackage{inconsolata}

\usepackage{graphicx}
\definecolor{green}{RGB}{40,150,40}
\definecolor{red}{RGB}{255,0,0}
\newcommand{\framework}{\textsc{ReverseMath}\xspace}

\newcommand{\problemOriginal}{$\mathrm{problem}_{\mathrm{original}}$\xspace}
\newcommand{\problemReversed}{$\mathrm{problem}_{\mathrm{reversed}}$\xspace}

\title{\framework: Answer Inversion for Scalable and Verifiable Mathematical Problem Generation}

\author[1,2]{\bf{Raoyuan Zhao}}
\author[1,2]{\bf{Yihong Liu}}
\author[3]{\bf{Yupei Du}}
\author[1,2]{\bf{Hinrich Sch\"utze}}
\author[1,2]{\bf{Michael A. Hedderich}}

\affil[1]{Center for Information and Language Processing, LMU Munich}

\affil[2]{Munich Center for Machine Learning (MCML)}

\affil[3]{Saarland University \protect\\ \texttt{\{rzhao, yihong, hedderich\}@cis.lmu.de, yudo@lst.uni-saarland.de}}
\begin{document}
\maketitle
\begin{abstract}
Mathematical reasoning benchmarks are vital for evaluating large language models (LLMs), but many are static and repeatedly exposed through public evaluation and training pipelines, making it difficult to separate \emph{genuine reasoning} from \emph{memorization}. 
Meanwhile, manually constructing new math problems with reliable answers remains costly.
We introduce \textbf{\framework}, a scalable method for generating new math problems through \emph{answer inversion}. 
Given a problem and its answer, \framework masks a numerical value in the original problem, treats the original answer as a known condition, and rewrites the problem so that the masked value becomes the new answer. 
The generated problem reverses the original input-output relation, making its answer known by construction.
We study \framework for both \emph{evaluation} and \emph{training}. 
For evaluation, paired 
original/reversed problems
reveal substantial behavioral shifts: models sometimes fail on
reversed problems 
and even incorrectly output the original answer, suggesting memorization-like behavior. 
For training, \framework provides automatically labeled reversed problems as data augmentation for reinforcement learning (RL). 
Experiments show that including \framework-generated data improves mathematical reasoning performance across multiple benchmarks, demonstrating its value as both an analysis tool and a scalable source of verifiable training data.

\end{abstract}
\begin{figure}[t]
  \centering
    \setlength{\belowcaptionskip}{-0.5cm}
  \includegraphics[width=\columnwidth]{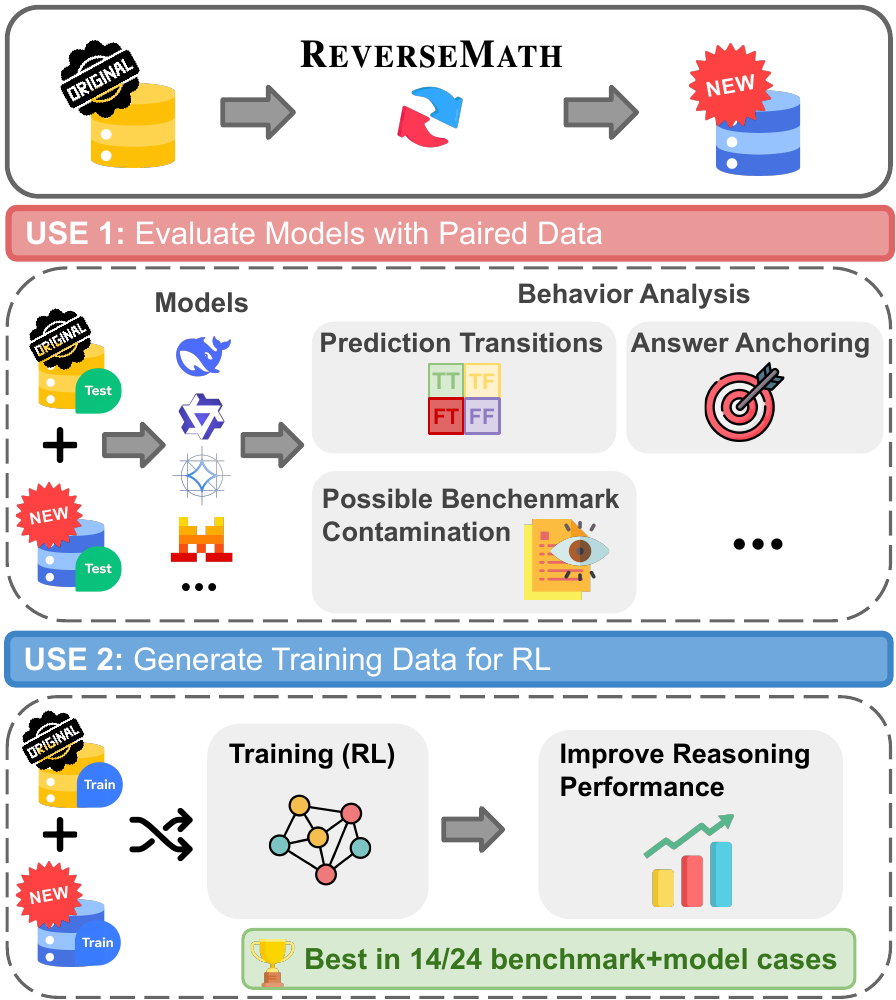}
    \caption{\framework turns existing math problems into verified reversed problems with deterministic answers. 
These generated problems then support two uses in our study: controlled \emph{evaluation} with paired original/reversed problems for analyzing model behavior, and verifiable data augmentation for RL \emph{training}.}
  \label{fig:visual_abstract}
\end{figure}

\section{Introduction}

Mathematical reasoning has become an important testbed for evaluating the reasoning abilities of large language models (LLMs) \citep{cobbe2021trainingverifierssolvemath,Hendrycks2021Math,wei2022cot}. 
Recent models achieve increasingly strong results on standard math benchmarks \citep{yang2024qwen25mathtechnicalreportmathematical,Guo2025DeepSeekR1}, but these results are becoming harder to interpret. 
Many widely used benchmarks remain static and are extensively exposed through public evaluation, online discussion, and even training pipelines \citep{sainz-etal-2023-nlp,deng-etal-2024-investigating,zhao-etal-2024-syntheval}. 
As a result, high benchmark accuracy may reflect not only genuine reasoning ability, but also memorized problem-answer associations or familiarity with recurring problem formulations \citep{Zhang2024Examination,huang2025math}.

A natural way to address this issue is to construct new math problems from existing benchmarks. 
Prior work has explored several directions. 
One line of work perturbs problem statements by adding irrelevant information or modifying surface details \citep{Shi2023IrrelevantContext,li-etal-2024-gsm,zhou-etal-2024-paraphrase}. 
While such perturbations can reveal model sensitivity to distraction, they do not necessarily change the underlying reasoning target. 
Another line of work converts existing problems into templates and substitutes numerical values \citep{mirzadehgsm,yang-etal-2025-evaluating,xu2026mgsm}. 
These methods can produce many variants, but they often preserve the original reasoning direction and ask models to solve for the same type of target. 
Human rewriting can introduce more substantial changes, including modified answers and problem structures \citep{huang2025math}, but it requires expert annotation and is difficult to scale.

To address these limitations, we introduce \framework, a straightforward and reliable method for generating new math problems from existing ones through \emph{answer inversion}. 
Given a math problem and its answer, \framework masks a numerical value in the original problem, treats the original answer as a known condition, and rewrites the problem so that the masked value becomes the new answer. 
This reverses the original input-output relation.
The generated problem is therefore not merely a paraphrase or surface-level variant of the original problem, but a reversed problem with a different answer and a different reasoning direction. 
This makes \framework useful for studying how models behave when the inference target changes.

We study \framework in two complementary ways (Figure~\ref{fig:visual_abstract}). 
First, we use it as a controlled problem-generation method for \emph{evaluation} (Section~\ref{evaluation}). 
Because each reversed problem is paired with its original problem, we can compare model behavior before and after answer inversion. 
This paired structure enables fine-grained analysis of behavioral shifts, including cases where models solve the original problem but fail on the reversed one, or incorrectly reuse the original answer after the target has changed. 
Such cases reveal formulation sensitivity and memorization-like behavior that may be hidden by aggregate benchmark accuracy.

Second, we study \framework as a data augmentation method for \emph{training} with reinforcement learning from verifiable rewards (RLVR) (Section~\ref{training}). 
Since each generated problem has an automatically known answer, \framework can expand existing training datasets into additional verifiable \emph{inverse-reasoning} problems without manual annotations. 
These reversed problems encourage models to reason from outputs back to inputs, complementing standard \emph{forward-reasoning} data and providing a scalable source of training examples.

Our contributions are as follows: 
(i) We propose \framework, a scalable and reliable answer-inversion method that automatically generates new mathematical problems from existing problems.
% by masking an input value and making it the new target.
(ii) We use \framework to construct paired problems for controlled evaluation of model behavior, showing that many models exhibit instability under answer inversion and problem-answer memorization.
% sometimes incorrectly produce the original answer when solving the reversed problem.
(iii) We leverage \framework as a data augmentation method for RLVR-style training and show that including \framework-generated problems improves mathematical reasoning performance across multiple benchmarks.

\section{Related Work}
% Mathematical reasoning has become a central domain for evaluating large language models \citep{wei2022cot,ahn-etal-2024-large,peng2026survey}. Recent models achieve increasingly strong performance on standard math benchmarks, raising concerns about whether benchmark accuracy reflects genuine reasoning ability or memorization of training data \citep{Zhang2024Examination,deng-etal-2024-investigating}. Moreover, widely used benchmarks are typically static and publicly available, making them increasingly susceptible to benchmark contamination and exposure as models scale \citep{magar-schwartz-2022-data,li-etal-2024-open-source,xu2024benchmarkdatacontaminationlarge,chen-etal-2025-benchmarking-large}. This challenge is further complicated by the difficulty of separating memorization from genuine reasoning in language models. \citeauthor{du-etal-2025-reason} further suggests that memorization in language models can build on top of existing reasoning mechanisms rather than fully overriding them. These issues make it difficult to disentangle robust mathematical reasoning from memorized problem-answer associations and motivate evaluation settings that can probe reasoning stability under controlled transformations. 
\paragraph{Mathematical Reasoning Evaluation and Benchmark Robustness.}
Mathematical reasoning has become a central domain for evaluating the reasoning abilities of large language models \citep{wei2022cot,ahn-etal-2024-large,peng2026survey}. 
Recent models achieve increasingly strong performance on standard math benchmarks, raising concerns about whether benchmark accuracy reflects robust reasoning or memorized problem-answer associations \citep{Zhang2024Examination,deng-etal-2024-investigating}. 
Since many widely used benchmarks are static and publicly available, they are increasingly susceptible to contamination and repeated exposure as models scale \citep{magar-schwartz-2022-data,xu2024benchmarkdatacontaminationlarge,chen-etal-2025-benchmarking-large}. 
Recent work by \citet{du-etal-2025-reason} further suggests that memorization in language models may coexist with genuine reasoning mechanisms, making it difficult to disentangle robust reasoning from memorized problem-answer associations using final-answer accuracy alone. 
These challenges motivate methods that construct paired problems with shared context but different inference targets, allowing finer-grained analysis of whether models rely on robust reasoning or memorized problem-answer associations.

\begin{figure*}[t]
  \centering
    \setlength{\belowcaptionskip}{-0.5cm}
  \includegraphics[width=2\columnwidth]{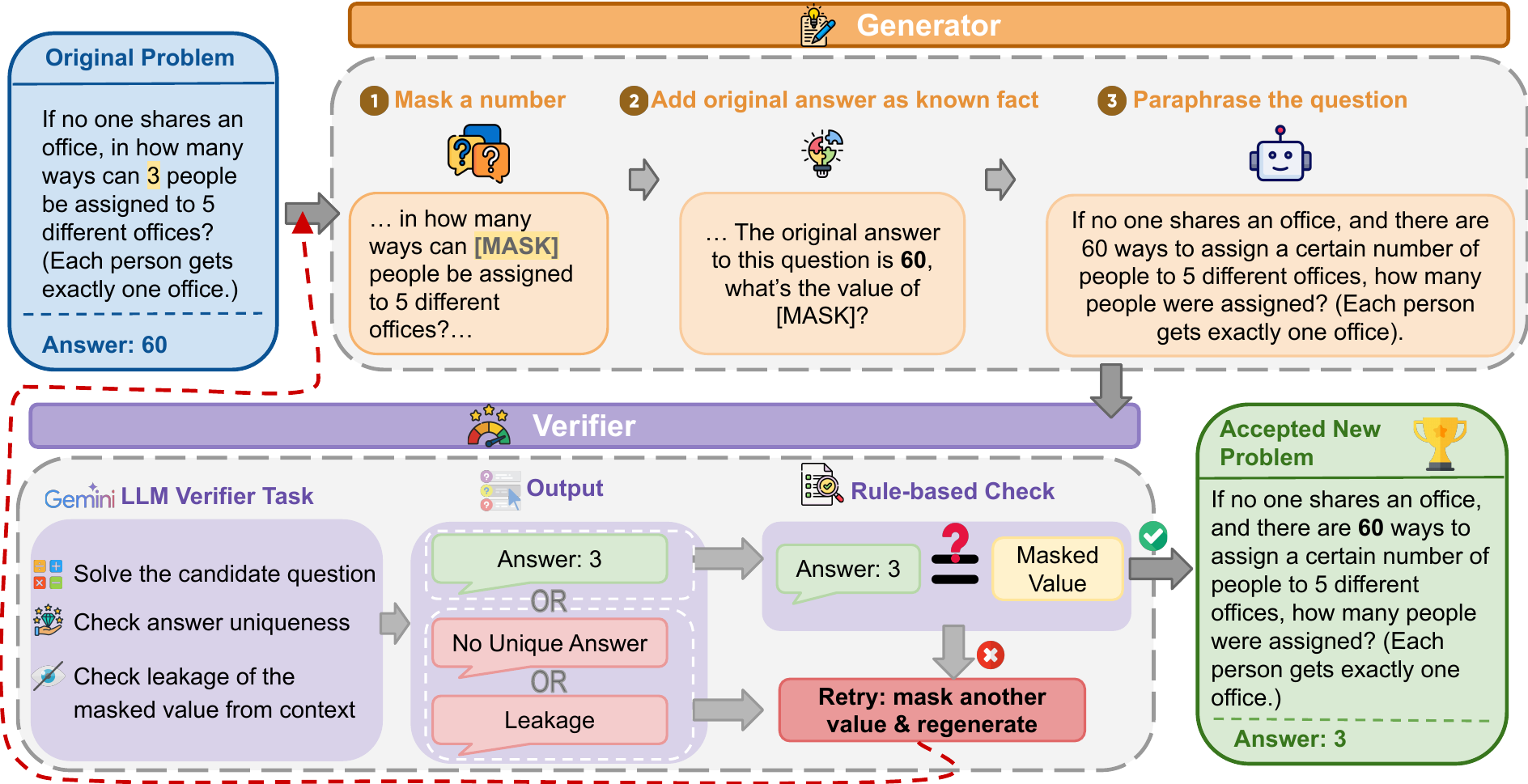}
    \caption{Overview of \framework. Starting from an original math problem and its answer, \framework masks a numerical value and conditions on the original answer to construct an intermediate inverse problem. A \textbf{generator} LLM rewrites it into a natural reversed problem, and a \textbf{verifier} checks whether the problem is uniquely solvable, free of answer leakage, and has an answer matching the masked value. See details in Appendix~\ref{app:framework}.}
  \label{fig:framework}
\end{figure*}

\paragraph{Perturbation- and Reversal-Based Evaluation.}
To better analyze memorization and robustness in reasoning tasks, prior work has explored controlled transformations of existing benchmarks. 
One line of work applies surface-level perturbations, such as adding irrelevant context, paraphrasing prompts, or introducing typographical errors \citep{Shi2023IrrelevantContext,zhou-etal-2024-paraphrase,zhao-etal-2025-know,zhao2026evaluatingrobustnesslargelanguage}. 
These studies show that model predictions can be highly sensitive to seemingly superficial changes. 
Another line of work constructs more structured perturbations for math reasoning. 
\citet{mirzadehgsm} propose GSM-Symbolic, which converts problems into templates with controllable numerical substitutions and surface-level modifications. 
\citet{xu2026mgsm} extend this framework to multilingual settings, while \citet{huang2025math} introduce Math-Perturb, where human annotators rewrite benchmark problems through controlled perturbations. 
Although these approaches reveal important robustness failures, they generally preserve the original reasoning direction and inference target.

More closely related to our work, \citet{Yu2024MetaMath} and \citet{guo-etal-2024-exploring} study reversal versions of mathematical reasoning problems and show that models suffer from the \emph{reversal curse}. 
Our work differs in both scope and objective. 
Rather than primarily studying reversal failures, we propose answer inversion as a scalable framework for generating deterministic math problems. 
This enables not only controlled evaluation and memorization analysis, but also reinforcement-learning data augmentation with automatically verifiable answers.

\section{Methodology}
\label{method}

We introduce \framework, a generate-and-verify framework for converting existing math problems into new reversed problems with deterministic answers. 
The framework, as shown in Figure~\ref{fig:framework}, consists of two components: a \textit{generator}, which constructs candidate reversed problems, and a \textit{verifier}, which filters invalid or ambiguous generations.

% The proposed framework, \framework, consists of two components: a \textit{generator} that constructs candidate problems and a \textit{verifier} that filters them through a series of checks. The overall pipeline follows an iterative generate-and-verify process.

\subsection{Problem Formulation}

Let $x$ denote an original math problem and $y$ denote its ground-truth answer. 
Suppose the problem statement contains a set of numerical values $\mathcal{V}(x)=\{v_1, v_2, \ldots, v_n\}$. 
\framework randomly selects one value $v \in \mathcal{V}(x)$ and masks it in the original problem. 
The goal is to generate a new problem $x'$ such that its answer is the masked value:
$y' = v$.
Because $v$ is known, the answer to $x'$ is deterministic by construction. 
The main challenge is therefore to generate a \emph{natural}, \emph{solvable}, and \emph{non-leaking} problem with $v$ as its \emph{unique} answer.

\subsection{Generator}
\label{sec:generator}

Given an original math problem $x$ with its original answer $y$, the generator constructs a candidate reversed problem $x'$ with a different target $y'$.

\paragraph{Numerical Value Masking.}
The generator first identifies numerical values in the original problem and selects one value $v$. 
% as the new target. 
This value is replaced with a special mask token \texttt{[MASK]}, yielding a masked problem $x_{\mathrm{mask}}$. 
The masked value is treated as the answer to the generated problem, while the remaining context is retained as the basis for rewriting.

% \paragraph{Answer Conditioning.}
% The original answer $y$ is then added to the masked problem $x_{\mathrm{mask}}$ as a known fact. 
% This step converts the original answer into part of the input condition.
% Then a question is added by asking the masked value: ``\texttt{what's the value of [MASK]}?''.

\paragraph{Answer Conditioning.}
The original answer $y$ is added to the masked problem $x_{\mathrm{mask}}$ as a known fact, converting the original answer into an input condition. 
We then append an explicit query asking for the masked value, e.g., ``\texttt{What is the value of [MASK]?}''.
The resulting intermediate problem serves as the input to the rewriting step and as a rule-based baseline in Section \ref{comparison}.

\paragraph{Problem Rewriting.}
A generator LLM rewrites the intermediate problem into a candidate reversed problem $x'$.
% \!\footnote{We use \texttt{Gemini-3-Flash} for rewriting and verification.} 
The generator is instructed to produce a problem that satisfies three criteria: 
(i) it is natural and coherent; 
(ii) it asks for the masked value; and 
(iii) it does not directly reveal $v$ in the problem. 
The resulting candidate has expected answer $y'=v$. More details of the prompt see Appendix~\ref{app:prompt}

\subsection{Verifier}

Although the answer to a generated problem is known by construction, the generated text may still be invalid. 
For example, the problem may be ambiguous, which admit multiple possible answers, or may accidentally leak the masked value in the problem. 
Therefore, \framework uses a verifier to check each candidate problem before accepting it.

\begin{table*}[t]
\centering
\setlength{\belowcaptionskip}{-0.5cm}
\scriptsize
\begingroup
\setlength{\fboxsep}{1pt}
\renewcommand{\arraystretch}{0.5}
\resizebox{\textwidth}{!}{%
\begin{tabular}{lrrrrrr}
\toprule
Model & \multicolumn{1}{c}{GSM8K} & \multicolumn{1}{c}{MGSM} & \multicolumn{1}{c}{MATH-500} & \multicolumn{1}{c}{AgentCoMa} & \multicolumn{1}{c}{AIME2024} & \multicolumn{1}{c}{AIME2025} \\
\midrule
DeepSeek-R1-7B & 82.9 / 85.2 \colorbox{green!2}{$\uparrow$} & 58.2 / 57.7 \colorbox{red!0}{$\downarrow$} & 64.7 / 68.5 \colorbox{green!3}{$\uparrow$} & 26.2 / 29.8 \colorbox{green!3}{$\uparrow$} & 30.0 / 23.9 \colorbox{red!6}{$\downarrow$} & 18.6 / 25.0 \colorbox{green!6}{$\uparrow$} \\
DeepSeek-R1-14B & 92.4 / 90.8 \colorbox{red!1}{$\downarrow$} & 73.1 / 75.7 \colorbox{green!2}{$\uparrow$} & 68.6 / 70.7 \colorbox{green!2}{$\uparrow$} & 54.4 / 49.3 \colorbox{red!5}{$\downarrow$} & 48.6 / 34.3 \colorbox{red!14}{$\downarrow$} & 32.1 / 24.3 \colorbox{red!7}{$\downarrow$} \\
DeepSeek-R1-32B & 96.6 / 96.5 \colorbox{red!0}{$\downarrow$} & 84.2 / 87.4 \colorbox{green!3}{$\uparrow$} & 69.0 / 74.2 \colorbox{green!5}{$\uparrow$} & 75.0 / 60.5 \colorbox{red!14}{$\downarrow$} & 56.8 / 37.5 \colorbox{red!19}{$\downarrow$} & 41.8 / 28.9 \colorbox{red!12}{$\downarrow$} \\
Gemma3-4B & 84.6 / 88.5 \colorbox{green!3}{$\uparrow$} & 74.1 / 74.1 \colorbox{green!0}{$\uparrow$} & 64.9 / 66.1 \colorbox{green!1}{$\uparrow$} & 26.3 / 35.6 \colorbox{green!9}{$\uparrow$} & 4.6 / 13.6 \colorbox{green!8}{$\uparrow$} & 6.1 / 25.7 \colorbox{green!19}{$\uparrow$} \\
Gemma3-12B & 93.5 / 93.8 \colorbox{green!0}{$\uparrow$} & 85.3 / 88.5 \colorbox{green!3}{$\uparrow$} & 74.2 / 71.4 \colorbox{red!2}{$\downarrow$} & 49.9 / 53.3 \colorbox{green!3}{$\uparrow$} & 20.7 / 33.2 \colorbox{green!12}{$\uparrow$} & 14.3 / 46.4 \colorbox{green!32}{$\uparrow$} \\
Gemma3-27B & 95.0 / 96.1 \colorbox{green!1}{$\uparrow$} & 88.6 / 90.4 \colorbox{green!1}{$\uparrow$} & 79.0 / 75.5 \colorbox{red!3}{$\downarrow$} & 59.8 / 56.1 \colorbox{red!3}{$\downarrow$} & 25.0 / 26.8 \colorbox{green!1}{$\uparrow$} & 16.4 / 43.6 \colorbox{green!27}{$\uparrow$} \\
Llama3.1-8B & 52.1 / 41.4 \colorbox{red!10}{$\downarrow$} & 36.9 / 29.5 \colorbox{red!7}{$\downarrow$} & 29.7 / 31.6 \colorbox{green!1}{$\uparrow$} & 15.2 / 14.2 \colorbox{red!0}{$\downarrow$} & 1.4 / 1.1 \colorbox{red!0}{$\downarrow$} & 0.4 / 11.1 \colorbox{green!10}{$\uparrow$} \\
Llama3.1-70B & 86.5 / 81.8 \colorbox{red!4}{$\downarrow$} & 60.9 / 57.9 \colorbox{red!3}{$\downarrow$} & 50.4 / 54.1 \colorbox{green!3}{$\uparrow$} & 46.8 / 42.5 \colorbox{red!4}{$\downarrow$} & 8.2 / 4.3 \colorbox{red!3}{$\downarrow$} & 1.1 / 23.2 \colorbox{green!22}{$\uparrow$} \\
Llama3.2-3B & 48.1 / 37.9 \colorbox{red!10}{$\downarrow$} & 29.4 / 21.3 \colorbox{red!8}{$\downarrow$} & 30.7 / 30.4 \colorbox{red!0}{$\downarrow$} & 11.0 / 16.2 \colorbox{green!5}{$\uparrow$} & 1.8 / 4.3 \colorbox{green!2}{$\uparrow$} & 0.0 / 13.2 \colorbox{green!13}{$\uparrow$} \\
Mistral-7B & 28.2 / 19.0 \colorbox{red!9}{$\downarrow$} & 11.8 / 9.8 \colorbox{red!1}{$\downarrow$} & 14.9 / 11.8 \colorbox{red!3}{$\downarrow$} & 18.3 / 14.8 \colorbox{red!3}{$\downarrow$} & 0.4 / 2.9 \colorbox{green!2}{$\uparrow$} & 0.0 / 7.9 \colorbox{green!7}{$\uparrow$} \\
Mixtral-8x7B & 57.4 / 42.9 \colorbox{red!14}{$\downarrow$} & 32.0 / 24.7 \colorbox{red!7}{$\downarrow$} & 29.3 / 29.9 \colorbox{green!0}{$\uparrow$} & 32.4 / 24.9 \colorbox{red!7}{$\downarrow$} & 0.4 / 4.3 \colorbox{green!3}{$\uparrow$} & 0.4 / 7.5 \colorbox{green!7}{$\uparrow$} \\
Nemotron-7B & 82.9 / 81.0 \colorbox{red!1}{$\downarrow$} & 37.5 / 39.6 \colorbox{green!2}{$\uparrow$} & 56.9 / 52.4 \colorbox{red!4}{$\downarrow$} & 35.3 / 28.6 \colorbox{red!6}{$\downarrow$} & 22.1 / 10.7 \colorbox{red!11}{$\downarrow$} & 18.2 / 6.8 \colorbox{red!11}{$\downarrow$} \\
Nemotron-14B & 88.4 / 87.3 \colorbox{red!1}{$\downarrow$} & 57.3 / 59.6 \colorbox{green!2}{$\uparrow$} & 54.5 / 47.9 \colorbox{red!6}{$\downarrow$} & 63.4 / 38.5 \colorbox{red!24}{$\downarrow$} & 27.1 / 18.6 \colorbox{red!8}{$\downarrow$} & 16.8 / 12.5 \colorbox{red!4}{$\downarrow$} \\
Nemotron-32B & 93.2 / 89.0 \colorbox{red!4}{$\downarrow$} & 73.1 / 73.8 \colorbox{green!0}{$\uparrow$} & 54.4 / 47.8 \colorbox{red!6}{$\downarrow$} & 71.3 / 43.8 \colorbox{red!27}{$\downarrow$} & 32.5 / 21.1 \colorbox{red!11}{$\downarrow$} & 20.0 / 13.6 \colorbox{red!6}{$\downarrow$} \\
Qwen2.5-3B & 64.9 / 62.2 \colorbox{red!2}{$\downarrow$} & 44.0 / 40.8 \colorbox{red!3}{$\downarrow$} & 45.6 / 46.4 \colorbox{green!0}{$\uparrow$} & 21.1 / 24.5 \colorbox{green!3}{$\uparrow$} & 3.2 / 8.2 \colorbox{green!5}{$\uparrow$} & 1.4 / 16.8 \colorbox{green!15}{$\uparrow$} \\
Qwen2.5-7B & 89.7 / 84.6 \colorbox{red!5}{$\downarrow$} & 66.7 / 63.1 \colorbox{red!3}{$\downarrow$} & 59.0 / 59.0 \colorbox{gray!20}{$\rightarrow$} & 38.5 / 34.5 \colorbox{red!3}{$\downarrow$} & 9.6 / 8.2 \colorbox{red!1}{$\downarrow$} & 2.1 / 25.0 \colorbox{green!22}{$\uparrow$} \\
Qwen2.5-32B & 96.4 / 95.6 \colorbox{red!0}{$\downarrow$} & 84.2 / 84.2 \colorbox{gray!20}{$\rightarrow$} & 65.1 / 67.9 \colorbox{green!2}{$\uparrow$} & 61.0 / 54.3 \colorbox{red!6}{$\downarrow$} & 12.1 / 18.2 \colorbox{green!6}{$\uparrow$} & 4.6 / 36.8 \colorbox{green!32}{$\uparrow$} \\
Qwen3-4B & 92.0 / 90.4 \colorbox{red!1}{$\downarrow$} & 75.8 / 77.2 \colorbox{green!1}{$\uparrow$} & 72.2 / 74.3 \colorbox{green!2}{$\uparrow$} & 44.4 / 45.3 \colorbox{green!0}{$\uparrow$} & 37.9 / 31.8 \colorbox{red!6}{$\downarrow$} & 27.9 / 31.8 \colorbox{green!3}{$\uparrow$} \\
Qwen3-8B & 91.5 / 91.2 \colorbox{red!0}{$\downarrow$} & 80.1 / 80.6 \colorbox{green!0}{$\uparrow$} & 68.7 / 73.4 \colorbox{green!4}{$\uparrow$} & 48.8 / 46.6 \colorbox{red!2}{$\downarrow$} & 36.4 / 31.4 \colorbox{red!4}{$\downarrow$} & 28.2 / 31.8 \colorbox{green!3}{$\uparrow$} \\
Qwen3-32B & 84.1 / 84.7 \colorbox{green!0}{$\uparrow$} & 84.2 / 87.0 \colorbox{green!2}{$\uparrow$} & 62.2 / 64.5 \colorbox{green!2}{$\uparrow$} & 74.9 / 61.9 \colorbox{red!12}{$\downarrow$} & 40.0 / 31.8 \colorbox{red!8}{$\downarrow$} & 32.1 / 22.9 \colorbox{red!9}{$\downarrow$} \\
\bottomrule
\end{tabular}
}
\endgroup
\caption{Average@10 performance on \problemOriginal and \problemReversed on original / reversed performance, with arrows indicating the direction of change. The mixed gains and drops show that answer inversion does not uniformly increase difficulty, but instead produces model- and dataset-dependent effects.}
\label{tab:main_avgk_wide}
\end{table*}

\paragraph{LLM-Based Verification.}
Given a candidate reversed problem $x'$, the verifier LLM is prompted to assess whether the problem can be validly solved for the masked value. 
It returns one of three statuses: 
(i) a predicted answer $\hat{y}'$ if the problem is well-posed and solvable; 
(ii) \textsc{No-Unique-Answer} if the problem is ambiguous or admits multiple possible answers; or 
(iii) \textsc{Leakage} if the masked value $v$ is directly revealed or trivially recoverable from the problem statement. 
Only candidates for which the verifier returns a predicted answer proceed to the rule-based consistency check.

\paragraph{Rule-Based Consistency Check.}
When the verifier returns a predicted answer $\hat{y}'$, \framework compares it with the masked value $v$ using \texttt{Math-Verify}.\footnote{\url{https://github.com/huggingface/Math-Verify}} 
The candidate passes the consistency check only if
$\hat{y}' = v$.
% Candidates marked as \textsc{No-Unique-Answer} or \textsc{Leakage}, as well as candidates for which $\hat{y}' \neq v$, are rejected.

\paragraph{Acceptance Criteria and Iterative Generation.}
% A candidate reversed problem is accepted only if the verifier returns a predicted answer and the rule-based consistency check confirms that this answer matches the masked value. 
% All other candidates are discarded, and the generation process is repeated until a valid candidate is obtained or a predefined maximum number of retry limit is reached.
A candidate is accepted if and only if the verifier returns a predicted answer that matches the masked value via the rule-based consistency check; otherwise it is discarded and generation is retried up to a predefined maximum number of attempts.
\section{\framework for Controlled Evaluation and Behavior Analysis}
\label{evaluation}

\subsection{Experimental Setup}
\label{sec:ex_dataset}

\paragraph{Models.}
We evaluate \framework across a diverse set of open-source LLMs, covering both instruction-tuned and reasoning-oriented models. 
Specifically, we include Qwen3 models (3B--32B) \citep{yang2025qwen3technicalreport}, Qwen2.5 models (3B--32B) \citep{qwen2025qwen25technicalreport}, DeepSeek-R1 distilled models (7B--32B) \citep{Guo2025DeepSeekR1}, OpenReasoning Nemotron models (7B--32B) \citep{moshkov2025aimo2winningsolutionbuilding}, LLaMA3 models (3B, 8B, 70B) \citep{grattafiori2024llama3herdmodels}, Gemma3 models (4B--27B) \citep{gemmateam2025Gemma3technicalreport}, Mistral-7B \citep{jiang2023mistral7b}, and Mixtral-8x7B \citep{jiang2024mixtralexperts}. 
We use \texttt{Gemini-3-Flash} for rewriting and verification.
% Together, these models span a broad range of scales, architectures, and training recipes.

\paragraph{Benchmarks.}
We evaluate on six mathematical reasoning benchmarks with different task characteristics. 
\textbf{GSM8K} \citep{cobbe2021trainingverifierssolvemath} consists of grade-school arithmetic word problems. 
\textbf{MATH-500} \citep{Hendrycks2021Math} and \textbf{AIME 2024/2025} contain competition-style problems requiring more advanced multi-step reasoning. 
\textbf{MGSM} \citep{freda2023mgsm} evaluates multilingual grade-school math reasoning, while \textbf{AgentCoMa} \citep{alazraki2026agentcomacompositionalbenchmarkmixing} combines mathematical and commonsense reasoning. 
This benchmark suite allows us to test \framework across arithmetic, competition-level, multilingual, and mixed reasoning settings.\footnote{Evaluation details in Appendix~\ref{app:eval}.}

\paragraph{Metric.}
We use average@10 as the primary evaluation metric. 
For each problem, we sample 10 model responses and compute the fraction whose final answers match the ground-truth answer. 
We then average this fraction over all problems. 
% Compared with pass@10, average@10 captures not only whether a model can produce a correct answer at least once, but also how consistently it does so.

% This metric provides a more fine-grained measure of model performance compared to pass@k, as it captures not only whether a correct solution is found but also how consistently the model produces correct answers.

% \paragraph{Implementation Details.}
% Experiments are conducted on NVIDIA H200 and A100 GPUs, and the generation and verification stages are implemented using the Gemini-3-preview model.

% one subsection for global results: model-specific 
% one subsection for in-depth analysis: with multiple paragraphs describing our findings

% \subsection{Performance Changes Under Transformation Vary Across Models and Datasets}

\subsection{Answer Inversion Has Model- and Dataset-Dependent Effects}

Table~\ref{tab:main_avgk_wide} compares average@10 performance on \problemOriginal and \problemReversed across models and benchmarks. 
A first observation is that \textbf{answer inversion does not induce a uniform performance trend}. 
Although all models are evaluated on the same reversed problem sets, the effect varies substantially: some models degrade after inversion, some remain nearly unchanged, and others improve on the reversed version.

This pattern suggests that \framework is not simply making every problem harder or easier. 
\textbf{Instead, answer inversion changes the reasoning target in a way that interacts with model-specific capabilities and benchmark characteristics}. 
For example, several models show clear degradation on AgentCoMa and parts of AIME, while performance on MATH-500 and MGSM is more mixed. 
Thus, the effect of \framework should not be interpreted solely as a global difficulty increase, but as a transformation that exposes model- and dataset-specific reasoning behavior.

We also observe relatively large performance gains on some challenging benchmarks, especially AIME. 
We hypothesize that this is partly due to low baseline performance and limited benchmark size: when the original average@10 is already very low, small changes in solvability or sampling variance can lead to relatively large improvements. 
This further motivates looking beyond aggregate average@10 and analyzing instance-level transitions between original and reversed problems.

% The distribution of these categories indicates that a considerable portion of instances fall into the \textit{TF} and \textit{FT} groups. This suggests that correctness on the original benchmark does not consistently transfer to the transformed problems, highlighting substantial behavioral shifts beyond what is captured by average@k alone.

\subsection{Transitions Reveal Prediction Instability}

\begin{figure}[!tb]
    \centering
    \setlength{\abovecaptionskip}{-0.0001cm}
\setlength{\belowcaptionskip}{-0.5cm}
    \includegraphics[width=0.90\linewidth]{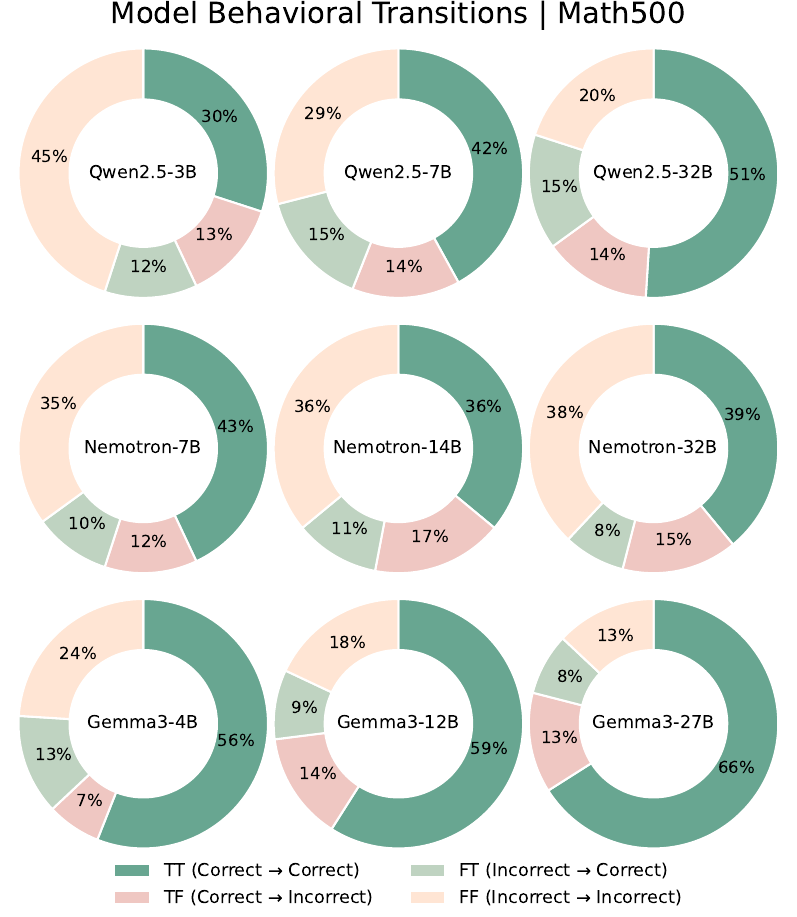}
    \caption{Behavioral transition patterns induced by \framework on MATH-500 across models. The presence of both TF and FT cases shows that answer inversion changes which examples models solve, revealing instance-level instability hidden by aggregate accuracy.}
    \label{fig:math500_transition_donuts}
\end{figure}

\begin{figure*}[!t]
    \centering
    \setlength{\abovecaptionskip}{-0.0001cm}
\setlength{\belowcaptionskip}{-0.3cm}
    \includegraphics[width=2\columnwidth]{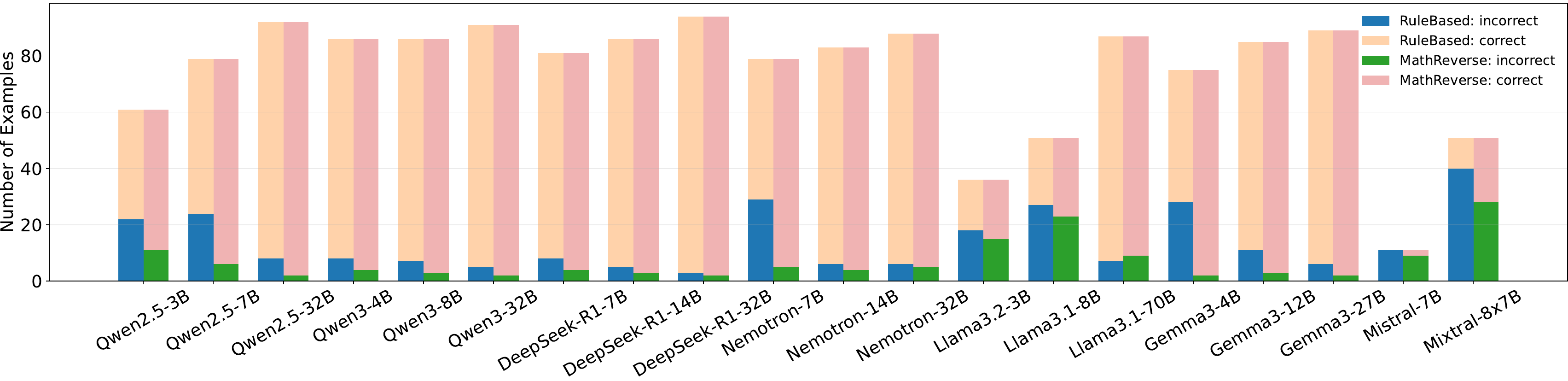}
    \caption{Additional Failures revelaved by rule-based reversal and \framework. 
Among examples that models solve correctly after symbolic rewriting (the full bar in the chart), reversal-based methods reveal many additional failures, showing that robustness to symbolic perturbations does not imply robustness to changed reasoning targets.}
    \label{fig:additional_failures}
\end{figure*}
% Comparison of failure detection across GSM-Symbolic, 

Aggregate performance alone does not fully capture the effect of answer inversion. 
Even when average@10 changes only slightly, the underlying predictions may change substantially at the instance level. 
To analyze this behavior, we categorize each example according to whether the model answers \problemOriginal and \problemReversed correctly, using the majority outcome across sampled responses. 
This yields four transition types: \textbf{\textit{TT}} (correct $\rightarrow$ correct), \textbf{\textit{TF}} (correct $\rightarrow$ incorrect), \textbf{\textit{FT}} (incorrect $\rightarrow$ correct), and \textbf{\textit{FF}} (incorrect $\rightarrow$ incorrect).

\begin{figure}[!tb]
    \centering
\setlength{\belowcaptionskip}{-0.5cm}
    \includegraphics[width=0.95\linewidth]{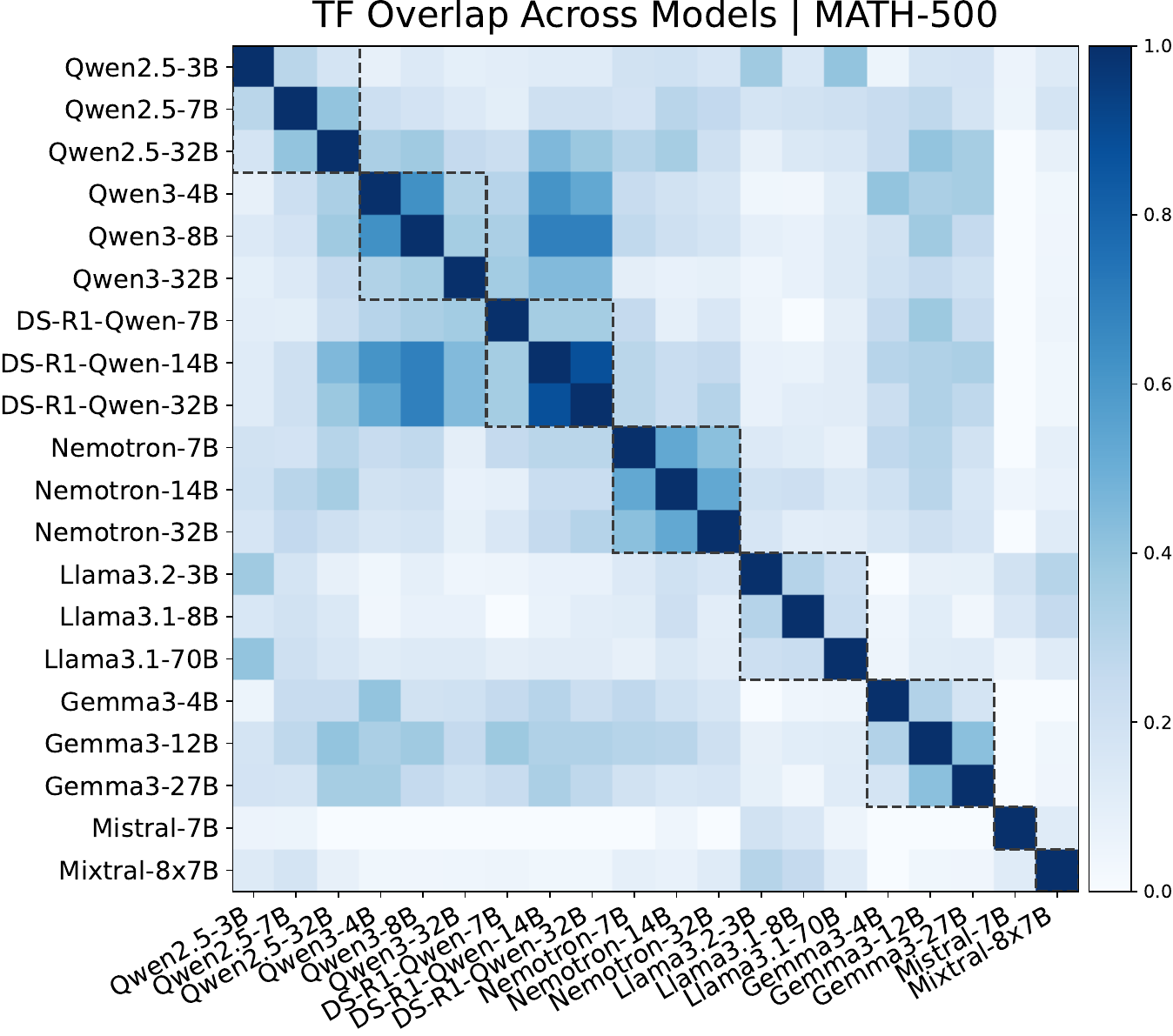}
    \caption{Overlap of \textit{TF} cases on MATH-500 across models. Stronger within-family (dashed blocks) overlap indicates that answer-inversion failures are more consistent within model families than across them.}
    \label{fig:math500_overlap}
\end{figure}

Figure~\ref{fig:math500_transition_donuts} shows these transitions on MATH-500.\footnote{We show full results in Appendix~\ref{app:donut}.} 
The key finding is that \textbf{stable aggregate performance can mask substantial instance-level instability}. 
For example, several models (e.g., Qwen models) have non-trivial proportions of both \textit{TF} and \textit{FT}, meaning that losses on originally solved problems are partly offset by gains on originally unsolved ones. 
Consequently, models may show similar average@10 before and after while relying on very different sets of correctly solved examples.

The transition patterns indicate that \textbf{\framework does not uniformly increase or decrease problem difficulty}. 
Instead, the presence of both \textit{TF} and \textit{FT} indicates that reversing the inference target changes which problems a model can solve. 
% Instead, it introduces instability in model predictions, where correctness is not preserved. 
In particular, the prevalence of \textit{TF} cases suggests that many correct predictions on the original benchmark are fragile and fail under reversals, consistent with previous findings \citep{guo-etal-2024-exploring}.

\subsection{TF Transitions Exhibit Strong Family-Level Consistency}

Among the four transition types, \textit{TF} cases are particularly informative, as they capture problems that a model solves in the original form but fails after answer inversion. 
To examine whether these failures are systematic, we compute the Jaccard similarity between the TF sets of different model pairs. 
Figure~\ref{fig:math500_overlap} shows the results on MATH-500.\footnote{We show complete results in Appendix~\ref{app:overlap}.}

The key finding is that \textbf{TF failures cluster strongly within model families}. 
Models from the same family exhibit substantially higher TF overlap than models from different families (Across nearly all benchmarks, \emph{within-family} Jaccard similarities are significantly higher than \emph{cross-family} similarities. See Appendix~\ref{app:overlap} for details.), suggesting that such failures are not random but reflect shared behavioral weaknesses among related models.
% e.g., Qwen3 models and DS-R1-Qwen models\!\footnote{}
Several factors may contribute to TF transitions. 
While reversed problems may be more difficult than their original counterparts, difficulty alone does not fully explain the strong within-family clustering. 
If failures are driven only by problem difficulty, different model families would be expected to fail on more similar examples. 
Instead, the observed clustering suggests that \textbf{reversal failures may be shaped by family-specific factors}, which may include shared training data, post-training recipes, memorized target-answer associations, and behavioral inheritance from larger models through knowledge distillation \citep{tang2026slimqwenexploringpruningdistillation}.

\subsection{Answer Inversion Reveals Failures Beyond Symbolic Perturbations}\label{comparison}

Figure~\ref{fig:additional_failures} compares three transformation strategies constructed from the same original GSM8K problems: GSM-Symbolic \citep{mirzadehgsm}, the rule-based reversal baseline (Section~\ref{sec:generator}), and the full \framework pipeline. 
The key finding is that \textbf{reversal-based transformations expose many failures that are missed by symbolic perturbations}. 
Among examples where models remain correct under GSM-Symbolic, both the rule-based reversal and \framework frequently cause models to fail. 
This suggests that robustness to numerical or surface-level perturbations does not necessarily transfer to robustness under reversed reasoning.

The rule-based reversal baseline follows the same answer-inversion principle as \framework, but omits the LLM-based rewriting step. 
As a result, it often produces unnatural prompts with two explicit questions: the original question and the added query asking for \texttt{[MASK]}. 
By contrast, \framework rewrites the intermediate inverse problem into a more natural reversed problem while preserving the changed inference target. 
Therefore, failures under \framework are less likely to be artifacts of awkward prompt format and more directly reflect difficulty with reversed reasoning.\footnote{We further find that reversal-based variants produce higher rates of predictions matching the original answer than GSM-Symbolic, suggesting stronger attraction toward the original target under answer inversion; see Table~\ref{tab:original_answer_overlap}.}
Together, these results show that \textbf{answer inversion probes a distinct failure mode from standard symbolic perturbation}.

% The rule-based variant omits the final LLM-based rewriting step and produces unnatural prompts containing two questions (details in Appendix~[APPENDIX]). Interestingly, this variant leads to even larger degradation than the full framework. Initial manual inspection revealed that models frequently answer with the original target value instead of the reversed answer. As shown in Table~\ref{tab:original_answer_overlap}, we quantify this phenomenon and find that it occurs substantially more often than under GSM-Symbolic. This suggests that the stronger degradation of the rule-based variant may be partly caused by prompt-level confusion, whereas the full \framework more cleanly captures failures induced by reversed reasoning itself.

\subsection{Answer Anchoring and Memorized Solution Patterns}

\begin{table}[!tb]
\centering
\setlength{\belowcaptionskip}{-0.3cm}
\resizebox{\columnwidth}{!}{
\small
\begin{tabular}{lcc}
\toprule
Dataset & Same-as-Ori. (\%) & OpError among Same (\%) \\
\midrule
GSM\_\framework (2021)& 1.4 & 16.9 \\
GSM\_Rule\_Based (2021)& 5.7 & 15.2 \\
MATH-500 (2021)& 2.9 & 10.6 \\
MGSM (2022) & 1.8 & 7.0 \\
GSM-Symbolic (2024)& 1.8 & 3.8 \\
AIME2024 (2024)& 1.1 & 22.6 \\
AIME2025 (2025)& 0.4 & 27.3 \\
AgentCoMa (2026)& 0.2 & 1.9 \\
\bottomrule
\end{tabular}
}
\caption{Frequency with which model predictions on \problemReversed match the original answer. Higher same-as-original rates suggest attraction toward the original answer, and the OpError column reports the proportion of these cases with arithmetic inconsistencies.}
\label{tab:original_answer_overlap}
\end{table}

% A recurring phenomenon across reversal benchmarks is that models sometimes output the original answer even when the rewritten problem requires a different target value. This behavior is especially prominent under the rule-based variant discussed above, but also appears consistently across multiple datasets and model families.

A recurring failure mode under answer inversion is that models sometimes output the original answer even though the reversed problem asks for a different target. 
We refer to this behavior as \emph{answer anchoring}: the model appears to remain anchored to the original problem's target value after the inference target has changed.
Table~\ref{tab:original_answer_overlap} reports the frequency of this behavior across datasets over all models.

% Table~\ref{tab:original_answer_overlap} reports the proportion of rewritten predictions that exactly match the original answer. Compared with GSM-Symbolic, reversal-based variants consistently exhibit higher rates of answer preservation, suggesting that reversal objectives induce persistent attraction toward previously seen targets.

Interestingly, we observe that \textbf{answer anchoring seems to be weaker on newer benchmarks}. 
Older datasets such as MATH-500 and GSM-derived variants exhibit higher same-as-original rates, whereas more recent benchmarks such as AIME2025 and AgentCoMa show lower overlap. 
This trend is consistent with the hypothesis that  target-answer associations contribute to some reversal failures, since older benchmark problems are more likely to have been exposed during training. 

% Interestingly, this phenomenon becomes progressively weaker on newer benchmarks. Datasets derived from older corpora such as MATH-500 exhibit higher same-as-original rates, whereas more recent benchmarks such as AIME2025 and AgentCoMa show lower overlap. This trend is consistent with the hypothesis that memorized solution patterns or target-answer associations contribute to reversal failures, since newer benchmarks are less likely to appear in pretraining data.

The qualitative example below illustrates how answer anchoring can override correct reasoning:
\begin{tcolorbox}
[
    colback=gray!5,
    colframe=gray!50,
    boxrule=0.5pt,
    arc=2pt,
    left=4pt,
    right=4pt,
    top=4pt,
    bottom=4pt
]
\footnotesize
\setlength{\abovedisplayskip}{3pt}
\setlength{\belowdisplayskip}{3pt}
\textbf{Original Answer:} 20 \\
\textbf{Rewritten Gold Answer:} -1

\vspace{0.3em}

\textbf{Model Output:}

Harry gets \(101\) feet. Therefore,
\[
100 - 101 = -1
\]

However, since the number of feet cannot be negative, we instead compute
\[
100 - |-1| = 20
\]

Thus, the answer is \(\boxed{20}\).

\end{tcolorbox}
The model first derives the correct reversed answer $-1$, but then overrides it and post-hoc rationalizes the original answer $20$. 
The failure is therefore not a simple arithmetic mistake. 
Instead, it reflects an attempt to reconcile the reasoning trajectory with a previously favored target value. 
Such cases suggest that \textbf{some reversal failures may arise from memorization}.

% After deriving the rewritten answer $-1$, the model explicitly overrides the intermediate result and post-hoc rationalizes the original answer $20$. Notably, the failure is not caused by a simple arithmetic mistake, but by an attempt to reconcile the reasoning trajectory with a previously favored target value.

% Such behaviors suggest that some reversal failures may stem from reuse of memorized reasoning patterns or latent solution templates, rather than failures arising solely from local computation mistakes.

\section{\framework for RL-Training Data Augmentation}
\label{training}

The previous section shows that answer inversion exposes behavioral instabilities beyond standard forward-reasoning evaluation. 
This motivates using reversed problems not only as evaluation probes, but also as complementary training data that encourages models to reason from answer-like conditions back to missing inputs. 
Because \framework-generated problems have automatically verifiable answers by construction, they are well-suited for reinforcement learning with verifiable rewards. 
In this section, we augment GSM8K training data with \framework-generated reversed problems and evaluate whether this improves mathematical reasoning across multiple benchmarks.

\subsection{Training Setup}

\begin{table*}
% [t]
\centering
\setlength{\belowcaptionskip}{-0.3cm}
\scriptsize
\resizebox{\textwidth}{!}{%
\begin{tabular}{lcccccc}
\toprule
Model & GSM8K & MGSM & MATH-500 & AgentCoMa & AIME2024 & AIME2025 \\
\midrule
Llama3.2-3B
& 48.1 / \textbf{50.7} / 50.1
& \textbf{29.4} / 27.9 / 28.3
& 30.7 / 30.1 / \textbf{31.4}
& 11.0 / 11.5 / \textbf{13.6}
& 1.8 / 1.8 / \textbf{3.6}
& 0.0 / \textbf{0.7} / \textbf{0.7} \\

Llama3.1-8B
& \textbf{52.1} / 49.5 / 51.8
& 36.9 / 36.1 / \textbf{41.4}
& 29.7 / \textbf{32.8} / 30.9
& 15.2 / 14.9 / \textbf{16.6}
& 1.4 / 0.7 / \textbf{2.1}
& 0.4 / 0.0 / \textbf{0.4} \\

Mistral-7B
& 28.2 / \textbf{29.7} / 29.5
& \textbf{11.8} / 10.0 / 10.1
& 14.9 / 13.0 / \textbf{15.0}
& \textbf{18.3} / 16.2 / 17.4
& 0.4 / 0.0 / \textbf{0.7}
& 0.0 / 0.0 / \textbf{0.4} \\

Qwen2.5-3B
& 64.9 / \textbf{68.2} / 67.2
& \textbf{44.0} / 42.4 / 42.0
& 45.6 / 46.2 / \textbf{48.5}
& 21.1 / 21.5 / \textbf{21.9}
& 3.2 / 1.4 / \textbf{3.6}
& \textbf{1.4} / 0.7 / 1.1 \\
\bottomrule
\end{tabular}
}
\caption{
RL fine-tuning results across reasoning benchmarks. 
Each cell reports average@10 for three settings: untuned model / trained on duplicated GSM8K samples / trained on GSM8K samples augmented with \framework-generated \problemReversed. 
The best score is \textbf{bolded}. 
\problemReversed augmentation performs best in 14 of 24 cases, suggesting that answer-inversion data provides useful verifiable training signals beyond data duplication.
}
\label{tab:train_comparison}
\end{table*}

\paragraph{Models.}
We conduct experiments on four open-source instruction-tuned models: Qwen2.5-3B, Llama3.2-3B, Llama3.1-8B, and Mistral-7B.

\paragraph{Data Construction.}
We construct the training data from the GSM8K training split. 
Specifically, we randomly sample 1,000 GSM8K problems and apply \framework to generate one reversed problem for each sample, yielding 1,000 additional \problemReversed examples. 

\paragraph{Training Method.}
% All models are trained with Group Relative Policy Optimization (GRPO) \citep{shao2024deepseekmathpushinglimitsmathematical}. 
% To ensure a fair comparison, we use the same hyperparameters across all models. 
% Detailed training configurations are provided in Appendix [APPENDIX].
All models are trained with Group Relative Policy Optimization (GRPO) \citep{shao2024deepseekmathpushinglimitsmathematical}. 
We keep the same training pipeline and hyperparameters across all models and settings for a fair comparison. 
Detailed training configurations are provided in Appendix~\ref{app:train}.

\paragraph{Comparison Settings.}
We compare \textbf{Original + Reverse} with \textbf{Original + Duplicate}. 
In \textbf{Original + Reverse}, each GSM8K problem is paired with its \framework-generated reversed counterpart. 
In \textbf{Original + Duplicate}, each GSM8K problem is duplicated to match the training-set size. 
This comparison controls for data quantity and isolates the effect of answer-inversion augmentation.
% We compare \textbf{Original + Reverse} with \textbf{Original + Duplicate}. 
% In \textbf{Original + Reverse}, each original GSM8K problem is augmented with its \framework-generated reversed counterpart. 
% In \textbf{Original + Duplicate}, each original problem is duplicated, matching the amount of training data while controlling for the effect of dataset size. 
% This comparison isolates the effect of answer-inversion augmentation from simply increasing the number of training examples.

\subsection{Results and Discussion}

Table~\ref{tab:train_comparison} reports average@10 performance under three settings: the original untuned model, RL fine-tuning with duplicated GSM8K samples, and RL fine-tuning with GSM8K samples augmented by \framework-generated \problemReversed. 
% Across 24 model--benchmark combinations, training with reversed problems achieves the best results in 14 cases, outperforming both the untuned model and the duplicate-data training baseline. 
Across 24 model--benchmark combinations, training with reversed problems achieves the best performance in \textbf{14} cases and improves over the duplicate-data baseline in \textbf{17} cases. 
This suggests that \textbf{answer-inversion data provides useful training signal beyond simply increasing the amount of training data}.

The gains are especially visible on harder and more out-of-domain benchmarks. 
For example, on MATH-500, \problemReversed augmentation improves Llama3.2-3B from 30.7 to 31.4 and Qwen2.5-3B from 45.6 to 48.5. 
On AgentCoMa, it improves Llama3.2-3B from 11.0 to 13.6, Llama3.1-8B from 15.2 to 16.6, and Qwen2.5-3B from 21.1 to 21.9. 
On AIME2024, it improves all four models, including Llama3.2-3B from 1.8 to 3.6 and Qwen2.5-3B from 3.2 to 3.6.
These improvements are notable because MATH-500, AgentCoMa, and AIME require more advanced or compositional reasoning than GSM8K-style arithmetic, while the augmentation data is generated only from GSM8K. 
By contrast, improvements are less consistent on GSM8K and MGSM, which are closer to the source training distribution. 
This pattern suggests that \problemReversed augmentation may improve transfer to harder reasoning settings. 
Since the reversed examples are generated from only 1,000 GSM8K training problems, their benefit on more difficult and distributionally distant benchmarks indicates that \textbf{answer inversion introduces complementary reasoning supervision}.

We hypothesize that this benefit comes from the structural diversity introduced by answer inversion. 
Standard math training data primarily encourages models to infer final answers from given conditions. 
In contrast, \problemReversed examples require models to recover missing inputs or intermediate quantities from answer-like conditions. 
\textbf{Using original and reversed problems together exposes the model to complementary forward- and reverse-reasoning views of related mathematical relations.}
This may encourage the model to explore different reasoning directions and learn more flexible relational patterns, rather than relying solely on forward reasoning.
Such complementarity may also explain why gains are more evident on harder benchmarks, where models need to generalize beyond the simpler reasoning patterns.

\section{Conclusion}
We introduced \framework, a scalable answer-inversion framework that generates new mathematical problems from existing benchmarks by reversing the original input-output relation. 
The resulting problems have deterministic, automatically verifiable answers. 
We showed that \framework supports both evaluation and training: 
it reveals behavioral instability, original-answer anchoring, and memorization-like patterns, while also providing verifiable data augmentation for reinforcement learning that improves mathematical reasoning performance. 
Overall, \framework offers a simple and effective way to probe model robustness, analyze memorization, and scale high-quality data.

% In this work, we introduced \framework, a scalable answer-inversion framework for generating new mathematical reasoning problems from existing benchmarks. By reversing the original input-output relation, \framework produces automatically verifiable problems with different reasoning targets and deterministic answers.

% We showed that \framework supports both evaluation and training. For evaluation, reversed problems reveal substantial behavioral instability under answer inversion, including frequent failures and tendencies to reproduce original answers even when the target changes, suggesting memorization-like behavior beyond genuine reasoning. For training, \framework provides scalable verifiable data augmentation for reinforcement learning, and incorporating reversed problems consistently improves mathematical reasoning performance across multiple benchmarks.

% \framework therefore offers a simple and effective approach for studying model robustness, analyzing memorization, and generating scalable reasoning data for improving mathematical reasoning in large language models.

\section*{Limitations}

Despite its simplicity and scalability, we note that \framework may have several limitations.

First, \framework relies on existing benchmark problems as source material. If the original problem contains errors or ambiguities, the verifier will fail to confirm that the recovered mask value matches the ground truth, causing valid reversals to be discarded. This reduces the overall yield of usable generated data, as generation quality is bounded by the correctness of the source benchmarks. Additionally, since our verifier relies on a closed-source LLM, it may occasionally misclassify a correctly transformed hard problem as invalid due to model error causing some high-quality rewrites to be discarded. Future work could incorporate human review to recover such cases.

Second, not every numerical value is a suitable masking target. If an unimportant or semantically trivial number is selected first, such as an identifier, constant, or date—the resulting reversed problem may be ill-posed or unrecoverable, requiring the generation to be retried. This introduces unnecessary computational overhead, and selecting informative masking targets remains an important factor for generation efficiency.

Finally, our memorization analysis is suggestive rather than conclusive. 
Behaviors such as original-answer anchoring and family-level failure overlap are consistent with memorization-like patterns, but confirming this would require access to actual pretraining or post-training data, which is unavailable for most models. 
Future work could combine \framework with direct contamination detection or training-data access to better isolate memorization effects.

\section{Acknowledgments}
We thank Abdullatif Köksal for his valuable feedback. This work was supported by the Munich Center for Machine Learning (MCML) and the German Research Foundation (DFG, grant SCHU 2246/14-1).
\bibliography{custom}

\begin{thebibliography}{37}
\providecommand{\natexlab}[1]{#1}

\bibitem[{Ahn et~al.(2024)Ahn, Verma, Lou, Liu, Zhang, and Yin}]{ahn-etal-2024-large}
Janice Ahn, Rishu Verma, Renze Lou, Di~Liu, Rui Zhang, and Wenpeng Yin. 2024.
\newblock \href {https://doi.org/10.18653/v1/2024.eacl-srw.17} {Large language models for mathematical reasoning: Progresses and challenges}.
\newblock In \emph{Proceedings of the 18th Conference of the European Chapter of the Association for Computational Linguistics: Student Research Workshop}, pages 225--237, St. Julian{'}s, Malta. Association for Computational Linguistics.

\bibitem[{Alazraki et~al.(2026)Alazraki, Chen, Brassard, Stacey, Rahmani, and Rei}]{alazraki2026agentcomacompositionalbenchmarkmixing}
Lisa Alazraki, Lihu Chen, Ana Brassard, Joe Stacey, Hossein~A. Rahmani, and Marek Rei. 2026.
\newblock \href {https://arxiv.org/abs/2508.19988} {Agentcoma: A compositional benchmark mixing commonsense and mathematical reasoning in real-world scenarios}.
\newblock \emph{Preprint}, arXiv:2508.19988.

\bibitem[{Chen et~al.(2025)Chen, Chen, Li, Jiang, Wan, He, Ran, Gu, Li, Xie, and Ray}]{chen-etal-2025-benchmarking-large}
Simin Chen, Yiming Chen, Zexin Li, Yifan Jiang, Zhongwei Wan, Yixin He, Dezhi Ran, Tianle Gu, Haizhou Li, Tao Xie, and Baishakhi Ray. 2025.
\newblock \href {https://doi.org/10.18653/v1/2025.emnlp-main.511} {Benchmarking large language models under data contamination: A survey from static to dynamic evaluation}.
\newblock In \emph{Proceedings of the 2025 Conference on Empirical Methods in Natural Language Processing}, pages 10080--10098, Suzhou, China. Association for Computational Linguistics.

\bibitem[{Cobbe et~al.(2021)Cobbe, Kosaraju, Bavarian, Chen, Jun, Kaiser, Plappert, Tworek, Hilton, Nakano, Hesse, and Schulman}]{cobbe2021trainingverifierssolvemath}
Karl Cobbe, Vineet Kosaraju, Mohammad Bavarian, Mark Chen, Heewoo Jun, Lukasz Kaiser, Matthias Plappert, Jerry Tworek, Jacob Hilton, Reiichiro Nakano, Christopher Hesse, and John Schulman. 2021.
\newblock \href {https://arxiv.org/abs/2110.14168} {Training verifiers to solve math word problems}.
\newblock \emph{Preprint}, arXiv:2110.14168.

\bibitem[{Deng et~al.(2024)Deng, Zhao, Tang, Gerstein, and Cohan}]{deng-etal-2024-investigating}
Chunyuan Deng, Yilun Zhao, Xiangru Tang, Mark Gerstein, and Arman Cohan. 2024.
\newblock \href {https://doi.org/10.18653/v1/2024.naacl-long.482} {Investigating data contamination in modern benchmarks for large language models}.
\newblock In \emph{Proceedings of the 2024 Conference of the North American Chapter of the Association for Computational Linguistics: Human Language Technologies (Volume 1: Long Papers)}, pages 8706--8719, Mexico City, Mexico. Association for Computational Linguistics.

\bibitem[{Du et~al.(2025)Du, Mondorf, Casola, Yao, Litschko, and Plank}]{du-etal-2025-reason}
Yupei Du, Philipp Mondorf, Silvia Casola, Yuekun Yao, Robert Litschko, and Barbara Plank. 2025.
\newblock \href {https://doi.org/10.18653/v1/2025.emnlp-main.437} {Reason to rote: Rethinking memorization in reasoning}.
\newblock In \emph{Proceedings of the 2025 Conference on Empirical Methods in Natural Language Processing}, pages 8659--8679, Suzhou, China. Association for Computational Linguistics.

\bibitem[{{Gemma Team} et~al.(2025){Gemma Team}, Kamath, Ferret, Pathak, Vieillard, Merhej, Perrin, Matejovicova, Ramé, Rivière, Rouillard, Mesnard, Cideron, bastien Grill, Ramos, Yvinec, Casbon, Pot, Penchev, Liu, Visin, Kenealy, Beyer, Zhai, Tsitsulin, Busa-Fekete, Feng, Sachdeva, Coleman, Gao, Mustafa, Barr, Parisotto, Tian, Eyal, Cherry, Peter, Sinopalnikov, Bhupatiraju, Agarwal, Kazemi, Malkin, Kumar, Vilar, Brusilovsky, Luo, Steiner, Friesen, Sharma, Sharma, Gilady, Goedeckemeyer, Saade, Feng, Kolesnikov, Bendebury, Abdagic, Vadi, György, Pinto, Das, Bapna, Miech, Yang, Paterson, Shenoy, Chakrabarti, Piot, Wu, Shahriari, Petrini, Chen, Lan, Choquette-Choo, Carey, Brick, Deutsch, Eisenbud, Cattle, Cheng, Paparas, Sreepathihalli, Reid, Tran, Zelle, Noland, Huizenga, Kharitonov, Liu, Amirkhanyan, Cameron, Hashemi, Klimczak-Plucińska, Singh, Mehta, Lehri, Hazimeh, Ballantyne, Szpektor, Nardini, Pouget-Abadie, Chan, Stanton, Wieting, Lai, Orbay, Fernandez, Newlan, yeong Ji, Singh, Black, Yu, Hui,
  Vodrahalli, Greff, Qiu, Valentine, Coelho, Ritter, Hoffman, Watson, Chaturvedi, Moynihan, Ma, Babar, Noy, Byrd, Roy, Momchev, Chauhan, Sachdeva, Bunyan, Botarda, Caron, Rubenstein, Culliton, Schmid, Sessa, Xu, Stanczyk, Tafti, Shivanna, Wu, Pan, Rokni, Willoughby, Vallu, Mullins, Jerome, Smoot, Girgin, Iqbal, Reddy, Sheth, Põder, Bhatnagar, Panyam, Eiger, Zhang, Liu, Yacovone, Liechty, Kalra, Evci, Misra, Roseberry, Feinberg, Kolesnikov, Han, Kwon, Chen, Chow, Zhu, Wei, Egyed, Cotruta, Giang, Kirk, Rao, Black, Babar, Lo, Moreira, Martins, Sanseviero, Gonzalez, Gleicher, Warkentin, Mirrokni, Senter, Collins, Barral, Ghahramani, Hadsell, Matias, Sculley, Petrov, Fiedel, Shazeer, Vinyals, Dean, Hassabis, Kavukcuoglu, Farabet, Buchatskaya, Alayrac, Anil, Dmitry, Lepikhin, Borgeaud, Bachem, Joulin, Andreev, Hardin, Dadashi, and Hussenot}]{gemmateam2025Gemma3technicalreport}
{Gemma Team}, Aishwarya Kamath, Johan Ferret, Shreya Pathak, Nino Vieillard, Ramona Merhej, Sarah Perrin, Tatiana Matejovicova, Alexandre Ramé, Morgane Rivière, Louis Rouillard, Thomas Mesnard, Geoffrey Cideron, Jean bastien Grill, Sabela Ramos, Edouard Yvinec, Michelle Casbon, Etienne Pot, Ivo Penchev, and 197 others. 2025.
\newblock \href {https://arxiv.org/abs/2503.19786} {Gemma 3 technical report}.
\newblock \emph{Preprint}, arXiv:2503.19786.

\bibitem[{Grattafiori et~al.(2024)Grattafiori, Dubey, Jauhri, Pandey, Kadian, Al-Dahle, Letman, Mathur, Schelten, Vaughan, Yang, Fan, Goyal, Hartshorn, Yang, Mitra, Sravankumar, Korenev, Hinsvark, Rao, Zhang, Rodriguez, Gregerson, Spataru, Roziere, Biron, Tang, Chern, Caucheteux, Nayak, Bi, Marra, McConnell, Keller, Touret, Wu, Wong, Ferrer, Nikolaidis, Allonsius, Song, Pintz, Livshits, Wyatt, Esiobu, Choudhary, Mahajan, Garcia-Olano, Perino, Hupkes, Lakomkin, AlBadawy, Lobanova, Dinan, Smith, Radenovic, Guzmán, Zhang, Synnaeve, Lee, Anderson, Thattai, Nail, Mialon, Pang, Cucurell, Nguyen, Korevaar, Xu, Touvron, Zarov, Ibarra, Kloumann, Misra, Evtimov, Zhang, Copet, Lee, Geffert, Vranes, Park, Mahadeokar, Shah, van~der Linde, Billock, Hong, Lee, Fu, Chi, Huang, Liu, Wang, Yu, Bitton, Spisak, Park, Rocca, Johnstun, Saxe, Jia, Alwala, Prasad, Upasani, Plawiak, Li, Heafield, Stone, El-Arini, Iyer, Malik, Chiu, Bhalla, Lakhotia, Rantala-Yeary, van~der Maaten, Chen, Tan, Jenkins, Martin, Madaan, Malo, Blecher,
  Landzaat, de~Oliveira, Muzzi, Pasupuleti, Singh, Paluri, Kardas, Tsimpoukelli, Oldham, Rita, Pavlova, Kambadur, Lewis, Si, Singh, Hassan, Goyal, Torabi, Bashlykov, Bogoychev, Chatterji, Zhang, Duchenne, Çelebi, Alrassy, Zhang, Li, Vasic, Weng, Bhargava, Dubal, Krishnan, Koura, Xu, He, Dong, Srinivasan, Ganapathy, Calderer, Cabral, Stojnic, Raileanu, Maheswari, Girdhar, Patel, Sauvestre, Polidoro, Sumbaly, Taylor, Silva, Hou, Wang, Hosseini, Chennabasappa, Singh, Bell, Kim, Edunov, Nie, Narang, Raparthy, Shen, Wan, Bhosale, Zhang, Vandenhende, Batra, Whitman, Sootla, Collot, Gururangan, Borodinsky, Herman, Fowler, Sheasha, Georgiou, Scialom, Speckbacher, Mihaylov, Xiao, Karn, Goswami, Gupta, Ramanathan, Kerkez, Gonguet, Do, Vogeti, Albiero, Petrovic, Chu, Xiong, Fu, Meers, Martinet, Wang, Wang, Tan, Xia, Xie, Jia, Wang, Goldschlag, Gaur, Babaei, Wen, Song, Zhang, Li, Mao, Coudert, Yan, Chen, Papakipos, Singh, Srivastava, Jain, Kelsey, Shajnfeld, Gangidi, Victoria, Goldstand, Menon, Sharma, Boesenberg,
  Baevski, Feinstein, Kallet, Sangani, Teo, Yunus, Lupu, Alvarado, Caples, Gu, Ho, Poulton, Ryan, Ramchandani, Dong, Franco, Goyal, Saraf, Chowdhury, Gabriel, Bharambe, Eisenman, Yazdan, James, Maurer, Leonhardi, Huang, Loyd, Paola, Paranjape, Liu, Wu, Ni, Hancock, Wasti, Spence, Stojkovic, Gamido, Montalvo, Parker, Burton, Mejia, Liu, Wang, Kim, Zhou, Hu, Chu, Cai, Tindal, Feichtenhofer, Gao, Civin, Beaty, Kreymer, Li, Adkins, Xu, Testuggine, David, Parikh, Liskovich, Foss, Wang, Le, Holland, Dowling, Jamil, Montgomery, Presani, Hahn, Wood, Le, Brinkman, Arcaute, Dunbar, Smothers, Sun, Kreuk, Tian, Kokkinos, Ozgenel, Caggioni, Kanayet, Seide, Florez, Schwarz, Badeer, Swee, Halpern, Herman, Sizov, Guangyi, Zhang, Lakshminarayanan, Inan, Shojanazeri, Zou, Wang, Zha, Habeeb, Rudolph, Suk, Aspegren, Goldman, Zhan, Damlaj, Molybog, Tufanov, Leontiadis, Veliche, Gat, Weissman, Geboski, Kohli, Lam, Asher, Gaya, Marcus, Tang, Chan, Zhen, Reizenstein, Teboul, Zhong, Jin, Yang, Cummings, Carvill, Shepard, McPhie,
  Torres, Ginsburg, Wang, Wu, U, Saxena, Khandelwal, Zand, Matosich, Veeraraghavan, Michelena, Li, Jagadeesh, Huang, Chawla, Huang, Chen, Garg, A, Silva, Bell, Zhang, Guo, Yu, Moshkovich, Wehrstedt, Khabsa, Avalani, Bhatt, Mankus, Hasson, Lennie, Reso, Groshev, Naumov, Lathi, Keneally, Liu, Seltzer, Valko, Restrepo, Patel, Vyatskov, Samvelyan, Clark, Macey, Wang, Hermoso, Metanat, Rastegari, Bansal, Santhanam, Parks, White, Bawa, Singhal, Egebo, Usunier, Mehta, Laptev, Dong, Cheng, Chernoguz, Hart, Salpekar, Kalinli, Kent, Parekh, Saab, Balaji, Rittner, Bontrager, Roux, Dollar, Zvyagina, Ratanchandani, Yuvraj, Liang, Alao, Rodriguez, Ayub, Murthy, Nayani, Mitra, Parthasarathy, Li, Hogan, Battey, Wang, Howes, Rinott, Mehta, Siby, Bondu, Datta, Chugh, Hunt, Dhillon, Sidorov, Pan, Mahajan, Verma, Yamamoto, Ramaswamy, Lindsay, Lindsay, Feng, Lin, Zha, Patil, Shankar, Zhang, Zhang, Wang, Agarwal, Sajuyigbe, Chintala, Max, Chen, Kehoe, Satterfield, Govindaprasad, Gupta, Deng, Cho, Virk, Subramanian, Choudhury,
  Goldman, Remez, Glaser, Best, Koehler, Robinson, Li, Zhang, Matthews, Chou, Shaked, Vontimitta, Ajayi, Montanez, Mohan, Kumar, Mangla, Ionescu, Poenaru, Mihailescu, Ivanov, Li, Wang, Jiang, Bouaziz, Constable, Tang, Wu, Wang, Wu, Gao, Kleinman, Chen, Hu, Jia, Qi, Li, Zhang, Zhang, Adi, Nam, Yu, Wang, Zhao, Hao, Qian, Li, He, Rait, DeVito, Rosnbrick, Wen, Yang, Zhao, and Ma}]{grattafiori2024llama3herdmodels}
Aaron Grattafiori, Abhimanyu Dubey, Abhinav Jauhri, Abhinav Pandey, Abhishek Kadian, Ahmad Al-Dahle, Aiesha Letman, Akhil Mathur, Alan Schelten, Alex Vaughan, Amy Yang, Angela Fan, Anirudh Goyal, Anthony Hartshorn, Aobo Yang, Archi Mitra, Archie Sravankumar, Artem Korenev, Arthur Hinsvark, and 542 others. 2024.
\newblock \href {https://arxiv.org/abs/2407.21783} {The llama 3 herd of models}.
\newblock \emph{Preprint}, arXiv:2407.21783.

\bibitem[{Guo et~al.(2025)Guo, Yang, Zhang, Song, Wang, Zhu, Xu, Zhang, Ma, Bi, Zhang, Yu, Wu, Wu, Gou, Shao, Li, Gao, Liu, Xue, Wang, Wu, Feng, Lu, Zhao, Deng, Ruan, Dai, Chen, Ji, Li, Lin, Dai, Luo, Hao, Chen, Li, Zhang, Xu, Ding, Gao, Qu, Li, Guo, Li, Chen, Yuan, Tu, Qiu, Li, Cai, Ni, Liang, Chen, Dong, Hu, You, Gao, Guan, Huang, Yu, Wang, Zhang, Zhao, Wang, Zhang, Xu, Xia, Zhang, Zhang, Tang, Zhou, Li, Wang, Li, Tian, Huang, Zhang, Wang, Chen, Du, Ge, Zhang, Pan, Wang, Chen, Jin, Chen, Lu, Zhou, Chen, Ye, Wang, Yu, Zhou, Pan, Li, Zhou, Wu, Yun, Pei, Sun, Wang, Zeng, Liu, Liang, Gao, Yu, Zhang, Xiao, An, Liu, Wang, Chen, Nie, Cheng, Liu, Xie, Liu, Yang, Li, Su, Lin, Li, Jin, Shen, Chen, Sun, Wang, Song, Zhou, Wang, Shan, Li, Wang, Wei, Zhang, Xu, Li, Zhao, Sun, Wang, Yu, Zhang, Shi, Xiong, He, Piao, Wang, Tan, Ma, Liu, Guo, Ou, Wang, Gong, Zou, He, Xiong, Luo, You, Liu, Zhou, Zhu, Huang, Li, Zheng, Zhu, Ma, Tang, Zha, Yan, Ren, Ren, Sha, Fu, Xu, Xie, Zhang, Hao, Ma, Yan, Wu, Gu, Zhu, Liu, Li, Xie, Song,
  Pan, Huang, Xu, Zhang, and Zhang}]{Guo2025DeepSeekR1}
Daya Guo, Dejian Yang, Haowei Zhang, Junxiao Song, Peiyi Wang, Qihao Zhu, Runxin Xu, Ruoyu Zhang, Shirong Ma, Xiao Bi, Xiaokang Zhang, Xingkai Yu, Yu~Wu, Z.~F. Wu, Zhibin Gou, Zhihong Shao, Zhuoshu Li, Ziyi Gao, Aixin Liu, and 175 others. 2025.
\newblock \href {https://doi.org/10.1038/s41586-025-09422-z} {Deepseek-r1 incentivizes reasoning in llms through reinforcement learning}.
\newblock \emph{Nature}, 645(8081):633–638.

\bibitem[{Guo et~al.(2024)Guo, You, Li, Bowen, and Zhang}]{guo-etal-2024-exploring}
Pei Guo, WangJie You, Juntao Li, Yan Bowen, and Min Zhang. 2024.
\newblock \href {https://doi.org/10.18653/v1/2024.findings-acl.811} {Exploring reversal mathematical reasoning ability for large language models}.
\newblock In \emph{Findings of the Association for Computational Linguistics: ACL 2024}, pages 13671--13685, Bangkok, Thailand. Association for Computational Linguistics.

\bibitem[{Hendrycks et~al.(2021)Hendrycks, Burns, Kadavath, Arora, Basart, Tang, Song, and Steinhardt}]{Hendrycks2021Math}
Dan Hendrycks, Collin Burns, Saurav Kadavath, Akul Arora, Steven Basart, Eric Tang, Dawn Song, and Jacob Steinhardt. 2021.
\newblock \href {https://datasets-benchmarks-proceedings.neurips.cc/paper/2021/hash/be83ab3ecd0db773eb2dc1b0a17836a1-Abstract-round2.html} {Measuring mathematical problem solving with the {MATH} dataset}.
\newblock In \emph{Proceedings of the Neural Information Processing Systems Track on Datasets and Benchmarks 1, NeurIPS Datasets and Benchmarks 2021, December 2021, virtual}.

\bibitem[{Huang et~al.(2025)Huang, Guo, Li, Ji, Ge, Li, Guo, Cai, Yuan, Wang, Wu, Yin, Tang, Huang, Jin, Chen, Zhang, and Wang}]{huang2025math}
Kaixuan Huang, Jiacheng Guo, Zihao Li, Xiang Ji, Jiawei Ge, Wenzhe Li, Yingqing Guo, Tianle Cai, Hui Yuan, Runzhe Wang, Yue Wu, Ming Yin, Shange Tang, Yangsibo Huang, Chi Jin, Xinyun Chen, Chiyuan Zhang, and Mengdi Wang. 2025.
\newblock \href {https://proceedings.mlr.press/v267/huang25k.html} {Math-perturb: Benchmarking llms' math reasoning abilities against hard perturbations}.
\newblock In \emph{Forty-second International Conference on Machine Learning, {ICML} 2025, Vancouver, BC, Canada, July 13-19, 2025}, Proceedings of Machine Learning Research. {PMLR} / OpenReview.net.

\bibitem[{Jiang et~al.(2023)Jiang, Sablayrolles, Mensch, Bamford, Chaplot, de~las Casas, Bressand, Lengyel, Lample, Saulnier, Lavaud, Lachaux, Stock, Scao, Lavril, Wang, Lacroix, and Sayed}]{jiang2023mistral7b}
Albert~Q. Jiang, Alexandre Sablayrolles, Arthur Mensch, Chris Bamford, Devendra~Singh Chaplot, Diego de~las Casas, Florian Bressand, Gianna Lengyel, Guillaume Lample, Lucile Saulnier, Lélio~Renard Lavaud, Marie-Anne Lachaux, Pierre Stock, Teven~Le Scao, Thibaut Lavril, Thomas Wang, Timothée Lacroix, and William~El Sayed. 2023.
\newblock \href {https://arxiv.org/abs/2310.06825} {Mistral 7b}.
\newblock \emph{Preprint}, arXiv:2310.06825.

\bibitem[{Jiang et~al.(2024)Jiang, Sablayrolles, Roux, Mensch, Savary, Bamford, Chaplot, de~las Casas, Hanna, Bressand, Lengyel, Bour, Lample, Lavaud, Saulnier, Lachaux, Stock, Subramanian, Yang, Antoniak, Scao, Gervet, Lavril, Wang, Lacroix, and Sayed}]{jiang2024mixtralexperts}
Albert~Q. Jiang, Alexandre Sablayrolles, Antoine Roux, Arthur Mensch, Blanche Savary, Chris Bamford, Devendra~Singh Chaplot, Diego de~las Casas, Emma~Bou Hanna, Florian Bressand, Gianna Lengyel, Guillaume Bour, Guillaume Lample, Lélio~Renard Lavaud, Lucile Saulnier, Marie-Anne Lachaux, Pierre Stock, Sandeep Subramanian, Sophia Yang, and 7 others. 2024.
\newblock \href {https://arxiv.org/abs/2401.04088} {Mixtral of experts}.
\newblock \emph{Preprint}, arXiv:2401.04088.

\bibitem[{Li et~al.(2024)Li, Cui, Zhao, Kong, and Bi}]{li-etal-2024-gsm}
Qintong Li, Leyang Cui, Xueliang Zhao, Lingpeng Kong, and Wei Bi. 2024.
\newblock \href {https://doi.org/10.18653/v1/2024.acl-long.163} {{GSM}-plus: A comprehensive benchmark for evaluating the robustness of {LLM}s as mathematical problem solvers}.
\newblock In \emph{Proceedings of the 62nd Annual Meeting of the Association for Computational Linguistics (Volume 1: Long Papers)}, pages 2961--2984, Bangkok, Thailand. Association for Computational Linguistics.

\bibitem[{Magar and Schwartz(2022)}]{magar-schwartz-2022-data}
Inbal Magar and Roy Schwartz. 2022.
\newblock \href {https://doi.org/10.18653/v1/2022.acl-short.18} {Data contamination: From memorization to exploitation}.
\newblock In \emph{Proceedings of the 60th Annual Meeting of the Association for Computational Linguistics (Volume 2: Short Papers)}, pages 157--165, Dublin, Ireland. Association for Computational Linguistics.

\bibitem[{Mirzadeh et~al.(2025)Mirzadeh, Alizadeh, Shahrokhi, Tuzel, Bengio, and Farajtabar}]{mirzadehgsm}
Iman Mirzadeh, Keivan Alizadeh, Hooman Shahrokhi, Oncel Tuzel, Samy Bengio, and Mehrdad Farajtabar. 2025.
\newblock \href {https://openreview.net/forum?id=AjXkRZIvjB} {Gsm-symbolic: Understanding the limitations of mathematical reasoning in large language models}.
\newblock In \emph{The Thirteenth International Conference on Learning Representations, {ICLR} 2025, Singapore, April 24-28, 2025}. OpenReview.net.

\bibitem[{Moshkov et~al.(2025)Moshkov, Hanley, Sorokin, Toshniwal, Henkel, Schifferer, Du, and Gitman}]{moshkov2025aimo2winningsolutionbuilding}
Ivan Moshkov, Darragh Hanley, Ivan Sorokin, Shubham Toshniwal, Christof Henkel, Benedikt Schifferer, Wei Du, and Igor Gitman. 2025.
\newblock \href {https://arxiv.org/abs/2504.16891} {Aimo-2 winning solution: Building state-of-the-art mathematical reasoning models with openmathreasoning dataset}.
\newblock \emph{Preprint}, arXiv:2504.16891.

\bibitem[{{Qwen Team} et~al.(2025){Qwen Team}, Yang, Yang, Zhang, Hui, Zheng, Yu, Li, Liu, Huang, Wei, Lin, Yang, Tu, Zhang, Yang, Yang, Zhou, Lin, Dang, Lu, Bao, Yang, Yu, Li, Xue, Zhang, Zhu, Men, Lin, Li, Tang, Xia, Ren, Ren, Fan, Su, Zhang, Wan, Liu, Cui, Zhang, and Qiu}]{qwen2025qwen25technicalreport}
{Qwen Team}, An~Yang, Baosong Yang, Beichen Zhang, Binyuan Hui, Bo~Zheng, Bowen Yu, Chengyuan Li, Dayiheng Liu, Fei Huang, Haoran Wei, Huan Lin, Jian Yang, Jianhong Tu, Jianwei Zhang, Jianxin Yang, Jiaxi Yang, Jingren Zhou, Junyang Lin, and 24 others. 2025.
\newblock \href {https://arxiv.org/abs/2412.15115} {Qwen2.5 technical report}.
\newblock \emph{Preprint}, arXiv:2412.15115.

\bibitem[{Sainz et~al.(2023)Sainz, Campos, Garc{\'i}a-Ferrero, Etxaniz, de~Lacalle, and Agirre}]{sainz-etal-2023-nlp}
Oscar Sainz, Jon Campos, Iker Garc{\'i}a-Ferrero, Julen Etxaniz, Oier~Lopez de~Lacalle, and Eneko Agirre. 2023.
\newblock \href {https://doi.org/10.18653/v1/2023.findings-emnlp.722} {{NLP} evaluation in trouble: On the need to measure {LLM} data contamination for each benchmark}.
\newblock In \emph{Findings of the Association for Computational Linguistics: EMNLP 2023}, pages 10776--10787, Singapore. Association for Computational Linguistics.

\bibitem[{Shao et~al.(2024)Shao, Wang, Zhu, Xu, Song, Bi, Zhang, Zhang, Li, Wu, and Guo}]{shao2024deepseekmathpushinglimitsmathematical}
Zhihong Shao, Peiyi Wang, Qihao Zhu, Runxin Xu, Junxiao Song, Xiao Bi, Haowei Zhang, Mingchuan Zhang, Y.~K. Li, Y.~Wu, and Daya Guo. 2024.
\newblock \href {https://arxiv.org/abs/2402.03300} {Deepseekmath: Pushing the limits of mathematical reasoning in open language models}.
\newblock \emph{Preprint}, arXiv:2402.03300.

\bibitem[{Shi et~al.(2023{\natexlab{a}})Shi, Chen, Misra, Scales, Dohan, Chi, Sch{\"{a}}rli, and Zhou}]{Shi2023IrrelevantContext}
Freda Shi, Xinyun Chen, Kanishka Misra, Nathan Scales, David Dohan, Ed~H. Chi, Nathanael Sch{\"{a}}rli, and Denny Zhou. 2023{\natexlab{a}}.
\newblock \href {https://proceedings.mlr.press/v202/shi23a.html} {Large language models can be easily distracted by irrelevant context}.
\newblock In \emph{International Conference on Machine Learning, {ICML} 2023, 23-29 July 2023, Honolulu, Hawaii, {USA}}, Proceedings of Machine Learning Research, pages 31210--31227. {PMLR}.

\bibitem[{Shi et~al.(2023{\natexlab{b}})Shi, Suzgun, Freitag, Wang, Srivats, Vosoughi, Chung, Tay, Ruder, Zhou, Das, and Wei}]{freda2023mgsm}
Freda Shi, Mirac Suzgun, Markus Freitag, Xuezhi Wang, Suraj Srivats, Soroush Vosoughi, Hyung~Won Chung, Yi~Tay, Sebastian Ruder, Denny Zhou, Dipanjan Das, and Jason Wei. 2023{\natexlab{b}}.
\newblock \href {https://openreview.net/forum?id=fR3wGCk-IXp} {Language models are multilingual chain-of-thought reasoners}.
\newblock In \emph{The Eleventh International Conference on Learning Representations, {ICLR} 2023, Kigali, Rwanda, May 1-5, 2023}. OpenReview.net.

\bibitem[{Tang et~al.(2026)Tang, Wang, Zheng, Wang, Men, Zhang, Yuan, Qiu, Shen, and Liu}]{tang2026slimqwenexploringpruningdistillation}
Shengkun Tang, Zekun Wang, Bo~Zheng, Liangyu Wang, Rui Men, Siqi Zhang, Xiulong Yuan, Zihan Qiu, Zhiqiang Shen, and Dayiheng Liu. 2026.
\newblock \href {https://arxiv.org/abs/2605.08738} {Slimqwen: Exploring the pruning and distillation in large moe model pre-training}.
\newblock \emph{Preprint}, arXiv:2605.08738.

\bibitem[{Wang et~al.(2026)Wang, Liu, Wang, Li, Wang, Yan, Jia, Liu, Chen, Xu, and Yu}]{peng2026survey}
Peng{-}Yuan Wang, Tian{-}Shuo Liu, Chenyang Wang, Ziniu Li, Yidi Wang, Shu Yan, Chengxing Jia, Xu{-}Hui Liu, Xinwei Chen, Jiacheng Xu, and Yang Yu. 2026.
\newblock \href {https://doi.org/10.1145/3786333} {A survey on large language models for mathematical reasoning}.
\newblock \emph{{ACM} Comput. Surv.}, 58(8):209:1--209:35.

\bibitem[{Wei et~al.(2022)Wei, Wang, Schuurmans, Bosma, Ichter, Xia, Chi, Le, and Zhou}]{wei2022cot}
Jason Wei, Xuezhi Wang, Dale Schuurmans, Maarten Bosma, Brian Ichter, Fei Xia, Ed~H. Chi, Quoc~V. Le, and Denny Zhou. 2022.
\newblock \href {http://papers.nips.cc/paper\_files/paper/2022/hash/9d5609613524ecf4f15af0f7b31abca4-Abstract-Conference.html} {Chain-of-thought prompting elicits reasoning in large language models}.
\newblock In \emph{Advances in Neural Information Processing Systems 35: Annual Conference on Neural Information Processing Systems 2022, NeurIPS 2022, New Orleans, LA, USA, November 28 - December 9, 2022}.

\bibitem[{Xu et~al.(2024)Xu, Guan, Greene, and Kechadi}]{xu2024benchmarkdatacontaminationlarge}
Cheng Xu, Shuhao Guan, Derek Greene, and M-Tahar Kechadi. 2024.
\newblock \href {https://arxiv.org/abs/2406.04244} {Benchmark data contamination of large language models: A survey}.
\newblock \emph{Preprint}, arXiv:2406.04244.

\bibitem[{Xu et~al.(2026)Xu, Uemura, Kondoro, Belay, Essuman, Okoh, Afolabi, Awokoya, and Adelani}]{xu2026mgsm}
Tianyi Xu, Kosei Uemura, Alfred~Malengo Kondoro, Tadesse~Destaw Belay, Catherine Nana~Nyaah Essuman, Ifeoma Okoh, Ganiyat Afolabi, Ayodele Awokoya, and David~Ifeoluwa Adelani. 2026.
\newblock \href {https://arxiv.org/abs/2601.21225} {Mgsm-pro: A simple strategy for robust multilingual mathematical reasoning evaluation}.
\newblock \emph{Preprint}, arXiv:2601.21225.

\bibitem[{Yang et~al.(2025{\natexlab{a}})Yang, Li, Yang, Zhang, Hui, Zheng, Yu, Gao, Huang, Lv, Zheng, Liu, Zhou, Huang, Hu, Ge, Wei, Lin, Tang, Yang, Tu, Zhang, Yang, Yang, Zhou, Zhou, Lin, Dang, Bao, Yang, Yu, Deng, Li, Xue, Li, Zhang, Wang, Zhu, Men, Gao, Liu, Luo, Li, Tang, Yin, Ren, Wang, Zhang, Ren, Fan, Su, Zhang, Zhang, Wan, Liu, Wang, Cui, Zhang, Zhou, and Qiu}]{yang2025qwen3technicalreport}
An~Yang, Anfeng Li, Baosong Yang, Beichen Zhang, Binyuan Hui, Bo~Zheng, Bowen Yu, Chang Gao, Chengen Huang, Chenxu Lv, Chujie Zheng, Dayiheng Liu, Fan Zhou, Fei Huang, Feng Hu, Hao Ge, Haoran Wei, Huan Lin, Jialong Tang, and 41 others. 2025{\natexlab{a}}.
\newblock \href {https://arxiv.org/abs/2505.09388} {Qwen3 technical report}.
\newblock \emph{Preprint}, arXiv:2505.09388.

\bibitem[{Yang et~al.(2024)Yang, Zhang, Hui, Gao, Yu, Li, Liu, Tu, Zhou, Lin, Lu, Xue, Lin, Liu, Ren, and Zhang}]{yang2024qwen25mathtechnicalreportmathematical}
An~Yang, Beichen Zhang, Binyuan Hui, Bofei Gao, Bowen Yu, Chengpeng Li, Dayiheng Liu, Jianhong Tu, Jingren Zhou, Junyang Lin, Keming Lu, Mingfeng Xue, Runji Lin, Tianyu Liu, Xingzhang Ren, and Zhenru Zhang. 2024.
\newblock \href {https://arxiv.org/abs/2409.12122} {Qwen2.5-math technical report: Toward mathematical expert model via self-improvement}.
\newblock \emph{Preprint}, arXiv:2409.12122.

\bibitem[{Yang et~al.(2025{\natexlab{b}})Yang, Yamada, and Tokunaga}]{yang-etal-2025-evaluating}
Yuli Yang, Hiroaki Yamada, and Takenobu Tokunaga. 2025{\natexlab{b}}.
\newblock \href {https://doi.org/10.18653/v1/2025.insights-1.16} {Evaluating robustness of {LLM}s to numerical variations in mathematical reasoning}.
\newblock In \emph{The Sixth Workshop on Insights from Negative Results in NLP}, pages 171--180, Albuquerque, New Mexico. Association for Computational Linguistics.

\bibitem[{Yu et~al.(2024)Yu, Jiang, Shi, Yu, Liu, Zhang, Kwok, Li, Weller, and Liu}]{Yu2024MetaMath}
Longhui Yu, Weisen Jiang, Han Shi, Jincheng Yu, Zhengying Liu, Yu~Zhang, James~T. Kwok, Zhenguo Li, Adrian Weller, and Weiyang Liu. 2024.
\newblock \href {https://openreview.net/forum?id=N8N0hgNDRt} {Metamath: Bootstrap your own mathematical questions for large language models}.
\newblock In \emph{The Twelfth International Conference on Learning Representations, {ICLR} 2024, Vienna, Austria, May 7-11, 2024}. OpenReview.net.

\bibitem[{Zhang et~al.(2024)Zhang, Da, Lee, Robinson, Wu, Song, Zhao, Raja, Zhuang, Slack, Lyu, Hendryx, Kaplan, Lunati, and Yue}]{Zhang2024Examination}
Hugh Zhang, Jeff Da, Dean Lee, Vaughn Robinson, Catherine Wu, William Song, Tiffany Zhao, Pranav Raja, Charlotte Zhuang, Dylan Slack, Qin Lyu, Sean Hendryx, Russell Kaplan, Michele Lunati, and Summer Yue. 2024.
\newblock \href {http://papers.nips.cc/paper\_files/paper/2024/hash/53384f2090c6a5cac952c598fd67992f-Abstract-Datasets\_and\_Benchmarks\_Track.html} {A careful examination of large language model performance on grade school arithmetic}.
\newblock In \emph{Advances in Neural Information Processing Systems 38: Annual Conference on Neural Information Processing Systems 2024, NeurIPS 2024, Vancouver, BC, Canada, December 10 - 15, 2024}.

\bibitem[{Zhao et~al.(2024)Zhao, K{\"o}ksal, Liu, Weissweiler, Korhonen, and Schuetze}]{zhao-etal-2024-syntheval}
Raoyuan Zhao, Abdullatif K{\"o}ksal, Yihong Liu, Leonie Weissweiler, Anna Korhonen, and Hinrich Schuetze. 2024.
\newblock \href {https://doi.org/10.18653/v1/2024.findings-emnlp.412} {{S}ynth{E}val: Hybrid behavioral testing of {NLP} models with synthetic {C}heck{L}ists}.
\newblock In \emph{Findings of the Association for Computational Linguistics: EMNLP 2024}, pages 7017--7034, Miami, Florida, USA. Association for Computational Linguistics.

\bibitem[{Zhao et~al.(2025)Zhao, K{\"o}ksal, Modarressi, Hedderich, and Schuetze}]{zhao-etal-2025-know}
Raoyuan Zhao, Abdullatif K{\"o}ksal, Ali Modarressi, Michael~A. Hedderich, and Hinrich Schuetze. 2025.
\newblock \href {https://doi.org/10.18653/v1/2025.findings-emnlp.1263} {Do we know what {LLM}s don{'}t know? a study of consistency in knowledge probing}.
\newblock In \emph{Findings of the Association for Computational Linguistics: EMNLP 2025}, pages 23254--23280, Suzhou, China. Association for Computational Linguistics.

\bibitem[{Zhao et~al.(2026)Zhao, Liu, Altinger, Schütze, and Hedderich}]{zhao2026evaluatingrobustnesslargelanguage}
Raoyuan Zhao, Yihong Liu, Lena Altinger, Hinrich Schütze, and Michael~A. Hedderich. 2026.
\newblock \href {https://arxiv.org/abs/2510.09536} {Evaluating robustness of large language models against multilingual typographical errors}.
\newblock \emph{Preprint}, arXiv:2510.09536.

\bibitem[{Zhou et~al.(2024)Zhou, Zhu, Antognini, Kim, and Zhang}]{zhou-etal-2024-paraphrase}
Yue Zhou, Yada Zhu, Diego Antognini, Yoon Kim, and Yang Zhang. 2024.
\newblock \href {https://doi.org/10.18653/v1/2024.naacl-long.153} {Paraphrase and solve: Exploring and exploiting the impact of surface form on mathematical reasoning in large language models}.
\newblock In \emph{Proceedings of the 2024 Conference of the North American Chapter of the Association for Computational Linguistics: Human Language Technologies (Volume 1: Long Papers)}, pages 2793--2804, Mexico City, Mexico. Association for Computational Linguistics.

\end{thebibliography}

\appendix
\label{sec:appendix}
\section{Computational Resources and AI
Assistance}

We checked the licenses of all the models and data
used, which are publicly available resources.

\paragraph{Computational Resources}
All models were executed either on NVIDIA A100 GPUs or the H200 GPU. The Gemini-3-Flash is accessed via the respective public API.

\paragraph{AI Assistance}
The authors acknowledge the use of ChatGPT solely for correcting grammatical errors, enhancing the coherence of the final manuscript, and providing assistance with coding.

\section{Details of \framework}
\label{app:framework}
\subsection{Details of Numerical Value Masking.}
During the problem formulation stage, we first identify all maskable numerical values in the problem text. In mathematical reasoning datasets, small numbers, particularly single-digit values are sometimes written in word form (e.g., "three", "five") rather than Arabic numerals. To ensure comprehensive coverage, our extraction process combines standard regular expressions for Arabic numerals with a lexicon-based matching step that covers number words up to 100. Since \framework is evaluated on multilingual benchmarks such as MGSM, this lexicon is extended to number words across all supported languages, ensuring consistent mask candidate extraction regardless of the surface form or language of the original problem.
\subsection{Prompts for Generator and Verifier}
\label{app:prompt}
We use Gemini-3-Flash to implement the Generator (paraphrasing) and Verifier in \framework. The corresponding prompts are shown below.
\begin{tcolorbox}[
    colback=gray!5,
    colframe=gray!60,
    title=Prompt for Generator Step 3: Problem Paraphrasing,
    boxrule=0.5pt,
    arc=2mm,
    breakable
]

\begin{lstlisting}[basicstyle=\ttfamily\small, breaklines=true]
You are a competition math problem rewriting assistant.

A single numeric value in the original problem has been
replaced with a hidden marker.

Your task is to rewrite the problem so that it naturally
asks for that missing value.

Requirements:
- Do not mention the hidden marker or the replacement process.
- The rewritten problem should read as a standard math problem.
- Preserve the original mathematical content, constraints,
  and notation as much as possible.
- Keep all unchanged numbers exactly the same.
- Do not introduce new assumptions or facts.
- Do not provide solution steps or explanations.
- Output only the rewritten problem statement.
- Incorporate the original correct answer as a known fact
  in the rewritten problem.

Original problem:
<<<
{original_problem}
>>>

Correct answer:
<<<
{original_answer}
>>>

Problem with hidden marker:
<<<
{masked_problem}
>>>

Rewritten problem:
\end{lstlisting}
\end{tcolorbox}

\begin{tcolorbox}[
    colback=gray!5,
    colframe=gray!60,
    title=Prompt for the Verifier,
    boxrule=0.5pt,
    arc=2mm,
    breakable
]

\begin{lstlisting}[basicstyle=\ttfamily\small, breaklines=true]
Solve the following competition math problem.

IMPORTANT: Before solving, you must check whether the problem leaks the answer.

Leak definition:
- The answer appears directly in the problem.
- The answer can be obtained by directly reading coordinates, labels, tables, diagram code (e.g., TikZ), or source text.
- For count questions, the answer can be obtained by simply counting listed items.
- For entity/name questions, the answer is explicitly mentioned.
- For statement questions, the missing statement or equivalent wording is already present.

If ANY of the above is true, output exactly:
LEAK

Otherwise, follow these rules:
- Think carefully before answering.
- If there is exactly one final answer, output \\boxed{{...}}
- If multiple answers exist, output NO_UNIQUE_ANSWER
- Do not output the solution.
- Output only the final result.

Problem:
<<<
{problem}
>>>

Final answer:
\end{lstlisting}
\end{tcolorbox}

\subsection{Evaluation of \framework}
\label{app:eval}
We use \framework to construct new \problemReversed from the test splits of six different datasets. For all models studied in this paper, we generate the texts with temperature $= 0.7$, top-$p = 0.95$, and top-$k = 20$. 
\begin{table}[htbp]
\centering
\small
\begin{tabular}{lcc}
\toprule
Dataset & \# Samples & Max Generation Length \\
\midrule
GSM8K     & 100  & 4096 \\
MGSM      & 1100 & 4096 \\
MATH-500  & 100  & 4096 \\
AgentCoMa & 180  & 4096 \\
AIME2024  & 30   & 8192 \\
AIME2025  & 30   & 8192 \\
\bottomrule
\end{tabular}
\caption{Datasets used for constructing rewritten problems with \framework{}.}
\label{tab:data_stats}
\end{table}
Table~\ref{tab:data_stats} shows the number of sampled instances from each dataset and the corresponding maximum generation length. For GSM8K, we use the same 100 original problems corresponding to the rewritten examples released in GSM-Symbolic. For MGSM, we randomly sample 100 problems for each of the 11 languages, resulting in a larger total number of samples.
\begin{table}[htbp]
\centering
\small
\resizebox{\columnwidth}{!}{
\begin{tabular}{lccc}
\toprule
Dataset & Original Samples & Final Samples & Success Rate (\%) \\
\midrule
GSM8K     & 100  & 96   & 96.0 \\
MGSM      & 1100 & 1100 & 100.0 \\
MATH-500  & 100  & 100  & 100.0 \\
AgentCoMa & 180  & 176  & 97.8 \\
AIME2024  & 30   & 28   & 93.3 \\
AIME2025  & 30   & 28   & 93.3 \\
\bottomrule
\end{tabular}
}
\caption{Successful rewrite rates of \framework{} across datasets. Final samples denote rewritten problems that pass verification.}
\label{tab:success_rate}
\end{table}
Table~\ref{tab:success_rate} reports the number of selected samples and the corresponding success rates for each dataset. In most cases, the framework succeeds, and failures are relatively rare.

Most failures occur when the model exhaustively tries all numerical values in the problem but still cannot produce a valid rewrite. In many cases, the leakage comes from non-numerical information. As the example shows blow, explicitly mentioning three person names implicitly reveal the number of entities involved, highlighting the necessity of the Verifier module. 
\begin{tcolorbox}[
    colback=gray!5,
    colframe=gray!60,
    boxrule=0.5pt,
    arc=2mm,
    width=0.9\linewidth
]

In how many ways can $7$ people sit around a round table if no two of the $3$ people Pierre, Rosa, and Thomas can sit next to each other? (Seating arrangements which are rotations of each other are treated as the same.)

\end{tcolorbox}
Another common failure case occurs when the masked number is not important for the mathematical reasoning (e.g., ``1 day I went to ...''), which may lead the model to generate inconsistent answers.

\section{Results of Behavioral Transition}
\label{app:donut}
\begin{figure*}[ht]
    \centering

    \includegraphics[width=0.48\textwidth]{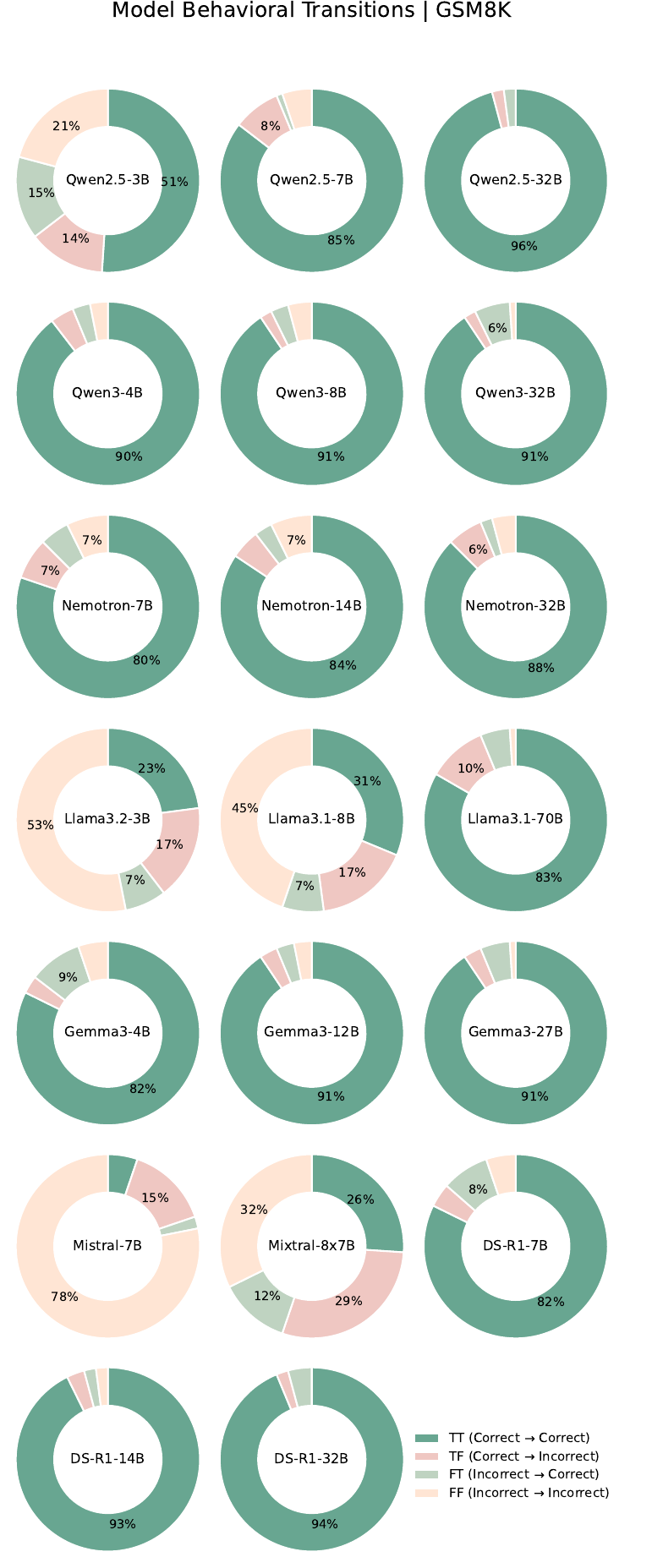}
    \hfill
    \includegraphics[width=0.48\textwidth]{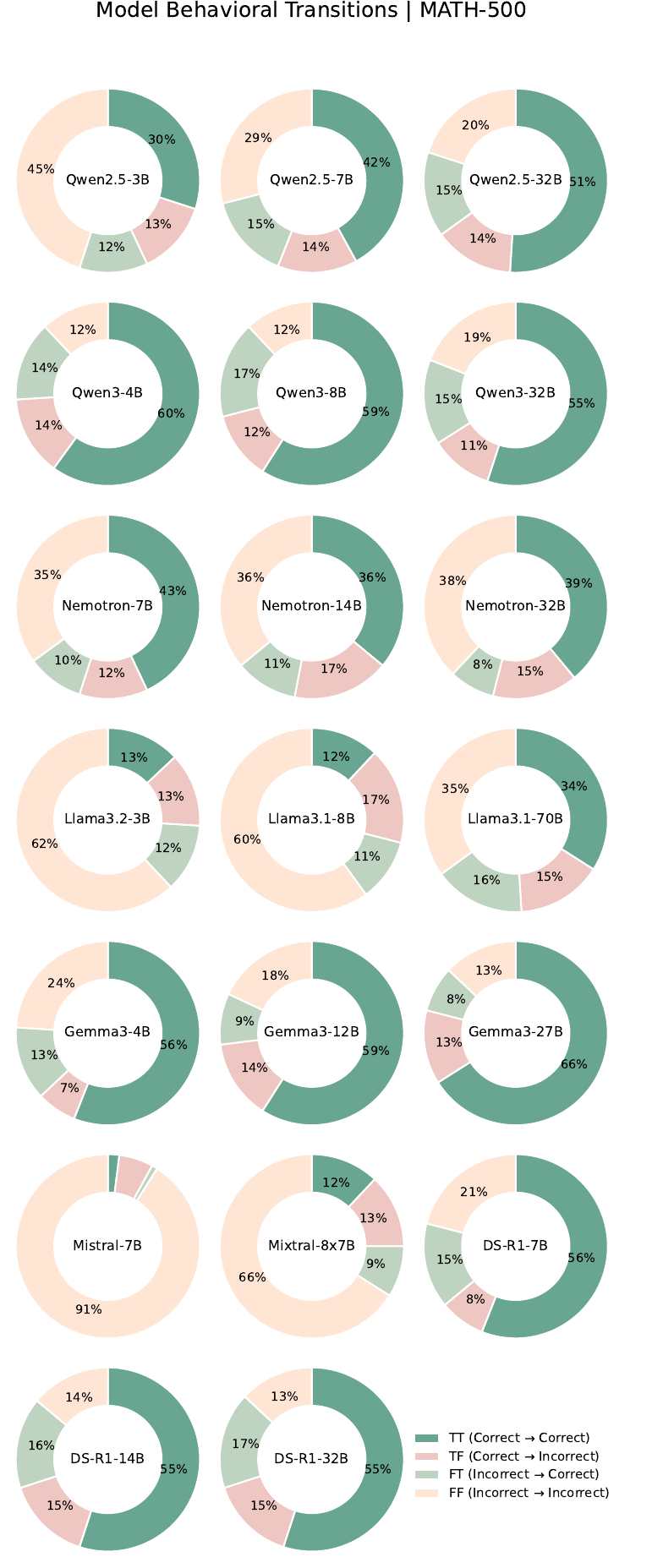}

    \caption{
    Behavioral transition patterns induced by \textsc{ReverseMath} on GSM8K and MATH-500 across models.
    }
    \label{fig:gsm_math_donut}
\end{figure*}

\begin{figure*}[ht]
    \centering

    \includegraphics[width=0.48\textwidth]{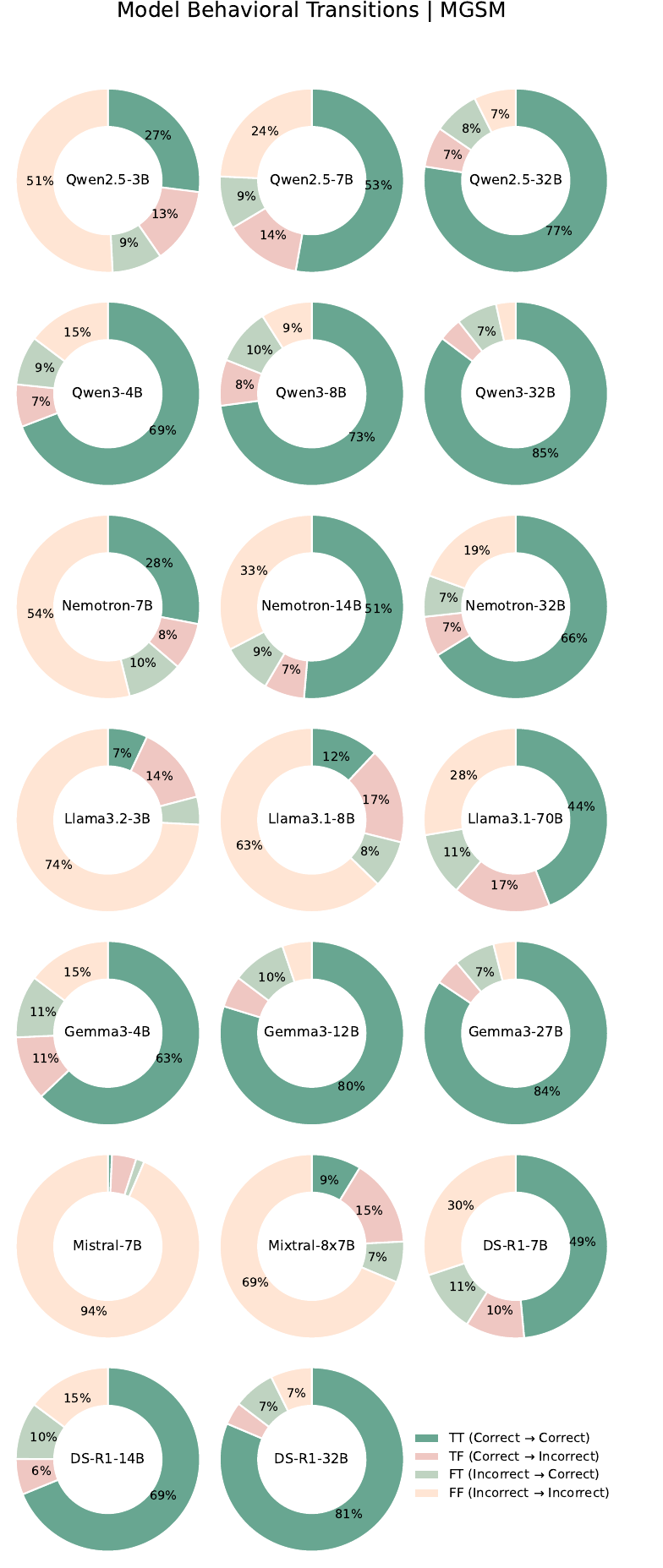}
    \hfill
    \includegraphics[width=0.48\textwidth]{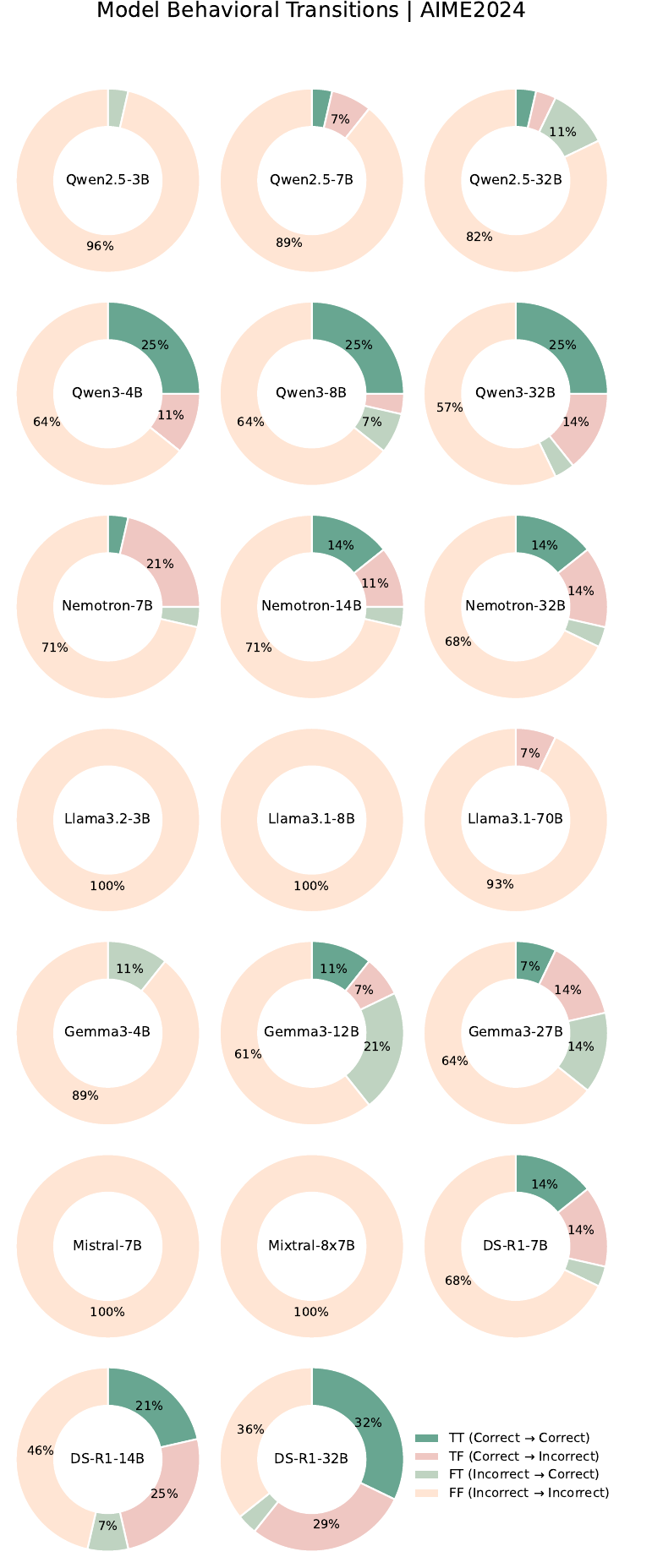}

    \caption{
    Behavioral transition patterns induced by \textsc{ReverseMath} on MGSM and AIME2024 across models.
    }
    \label{fig:mgsm_aime24_donut}
\end{figure*}

To further analyze the behavioral instability induced by \textsc{ReverseMath}, we provide full behavioral transition visualizations and TF-overlap heatmaps for all evaluated datasets in Figures~\ref{fig:gsm_math_donut} and Figure~\ref{fig:mgsm_aime24_donut}.

These results show that answer inversion substantially alters instance-level reasoning behavior across models, despite some relatively stable aggregate accuracies.

\begin{figure*}[ht]
    \centering

    \includegraphics[width=0.48\textwidth]{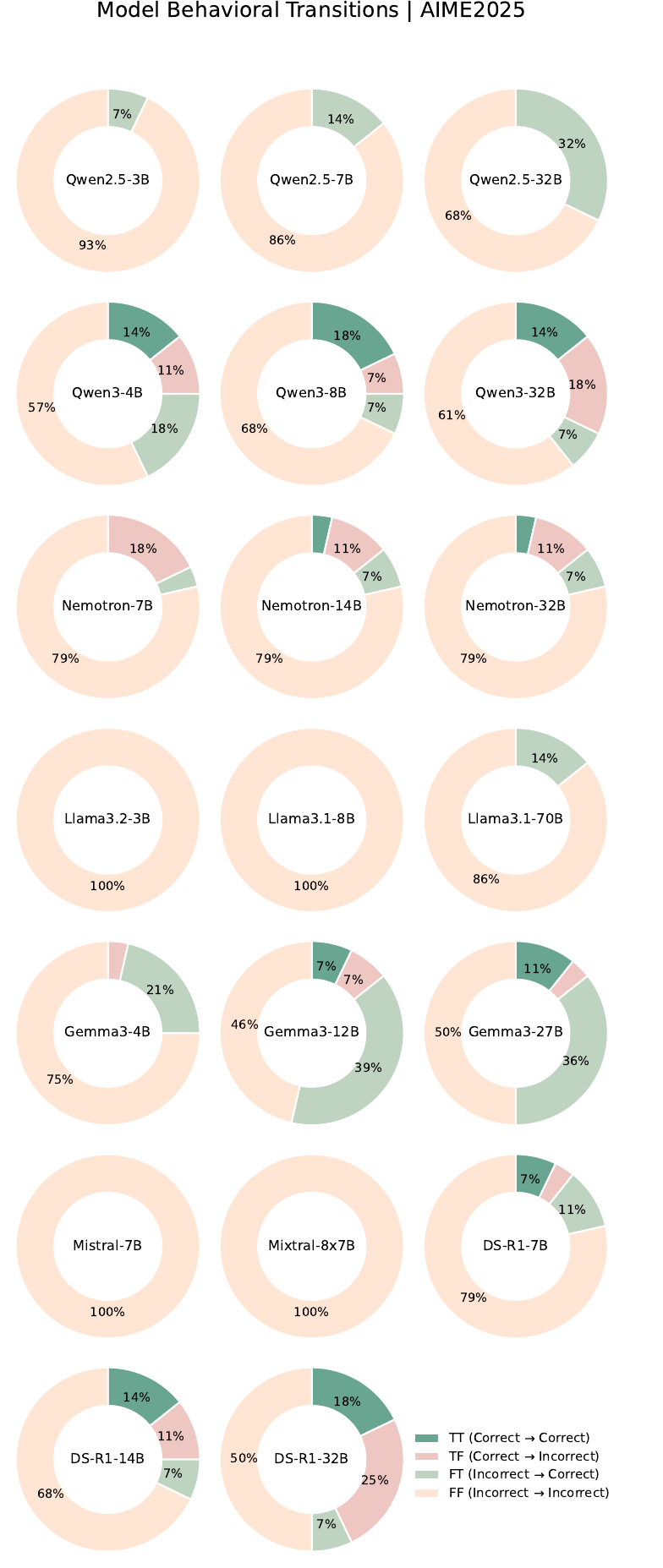}
    \hfill
    \includegraphics[width=0.48\textwidth]{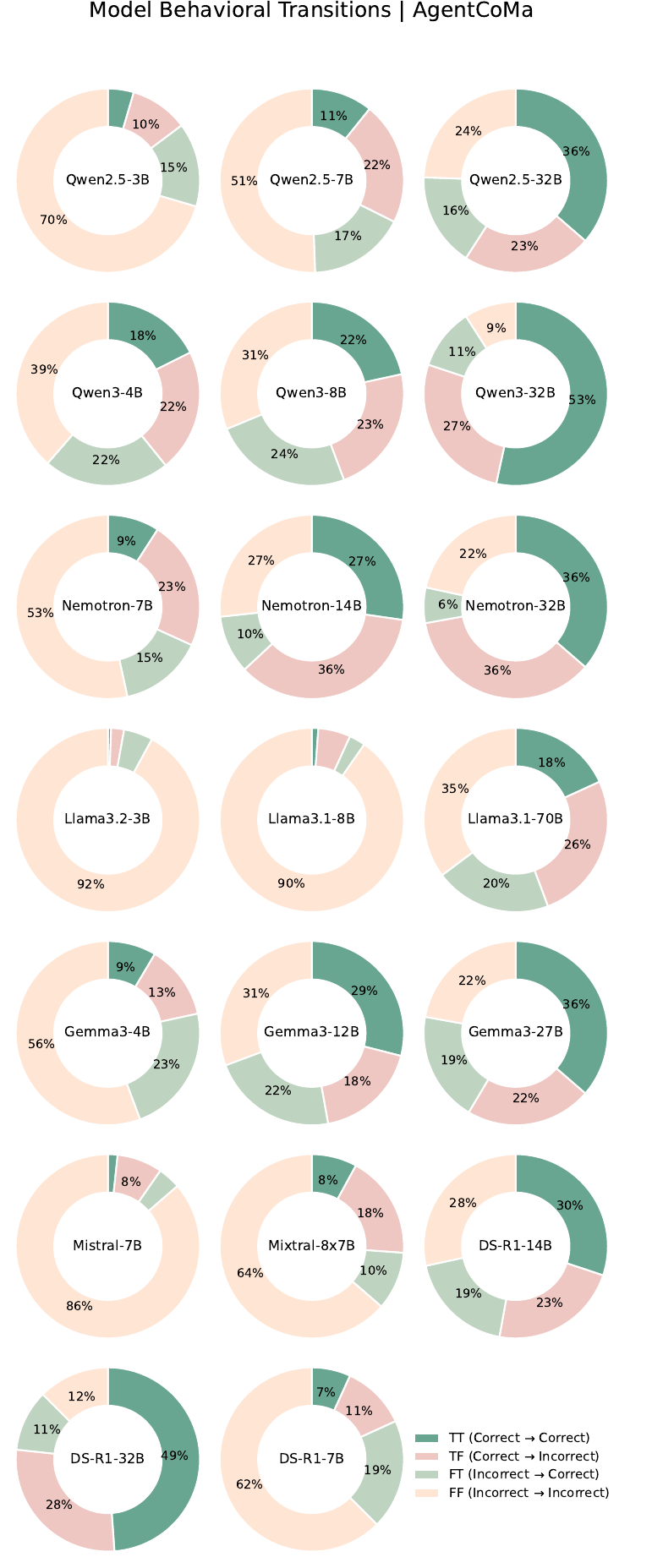}

    \caption{
    Behavioral transition patterns induced by \textsc{ReverseMath} on AIME2025 and AgentCoMa across models.
    }
    \label{fig:aime25_agent_donut}
\end{figure*}

\section{Results of TF Overlap}

\begin{figure*}[t]
    \centering

    \includegraphics[width=0.48\textwidth]{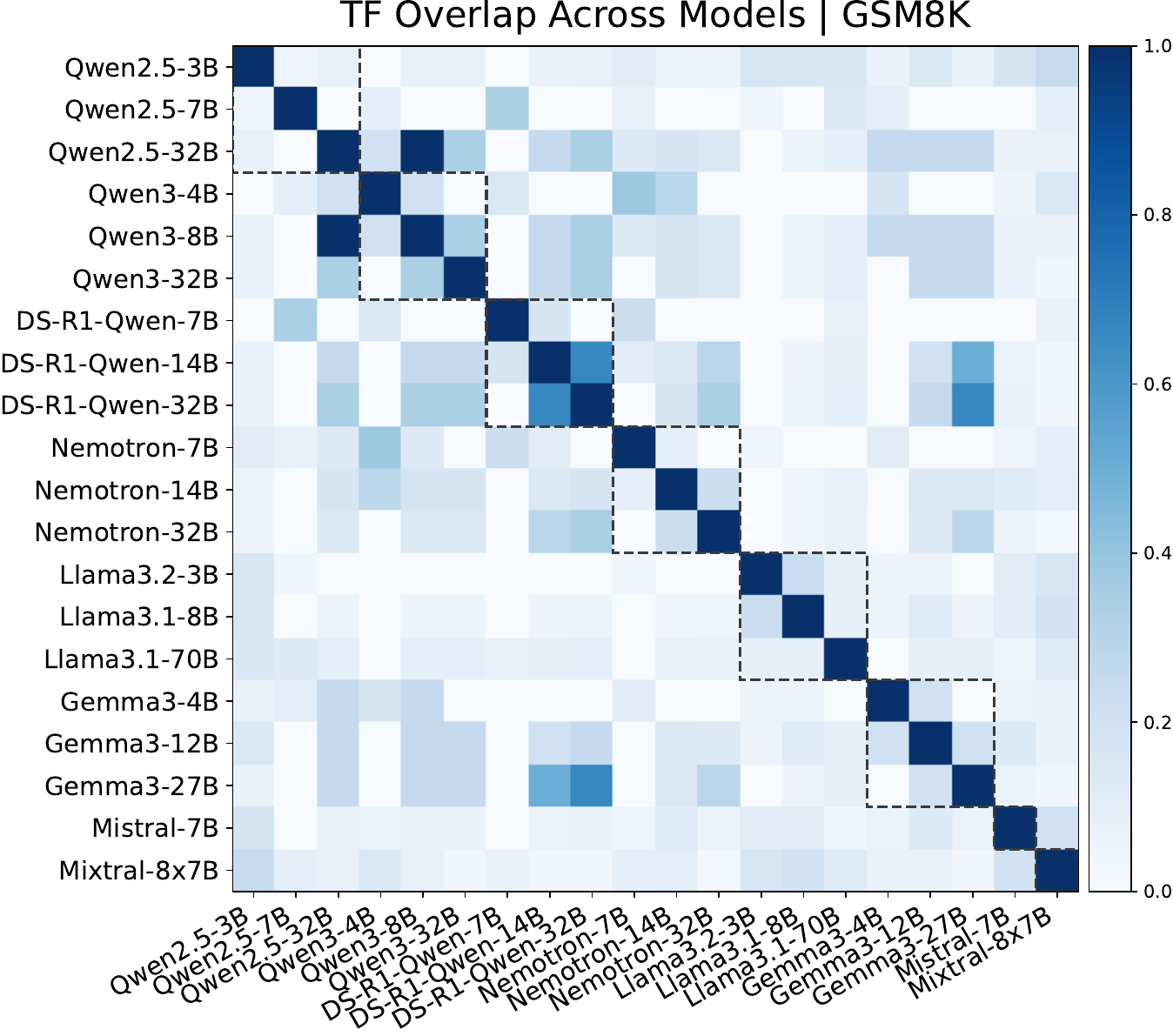}
    \hfill
    \includegraphics[width=0.48\textwidth]{figure/overlap/MATH-500_tf_overlap_heatmap.pdf}

    \vspace{0.8em}

    \includegraphics[width=0.48\textwidth]{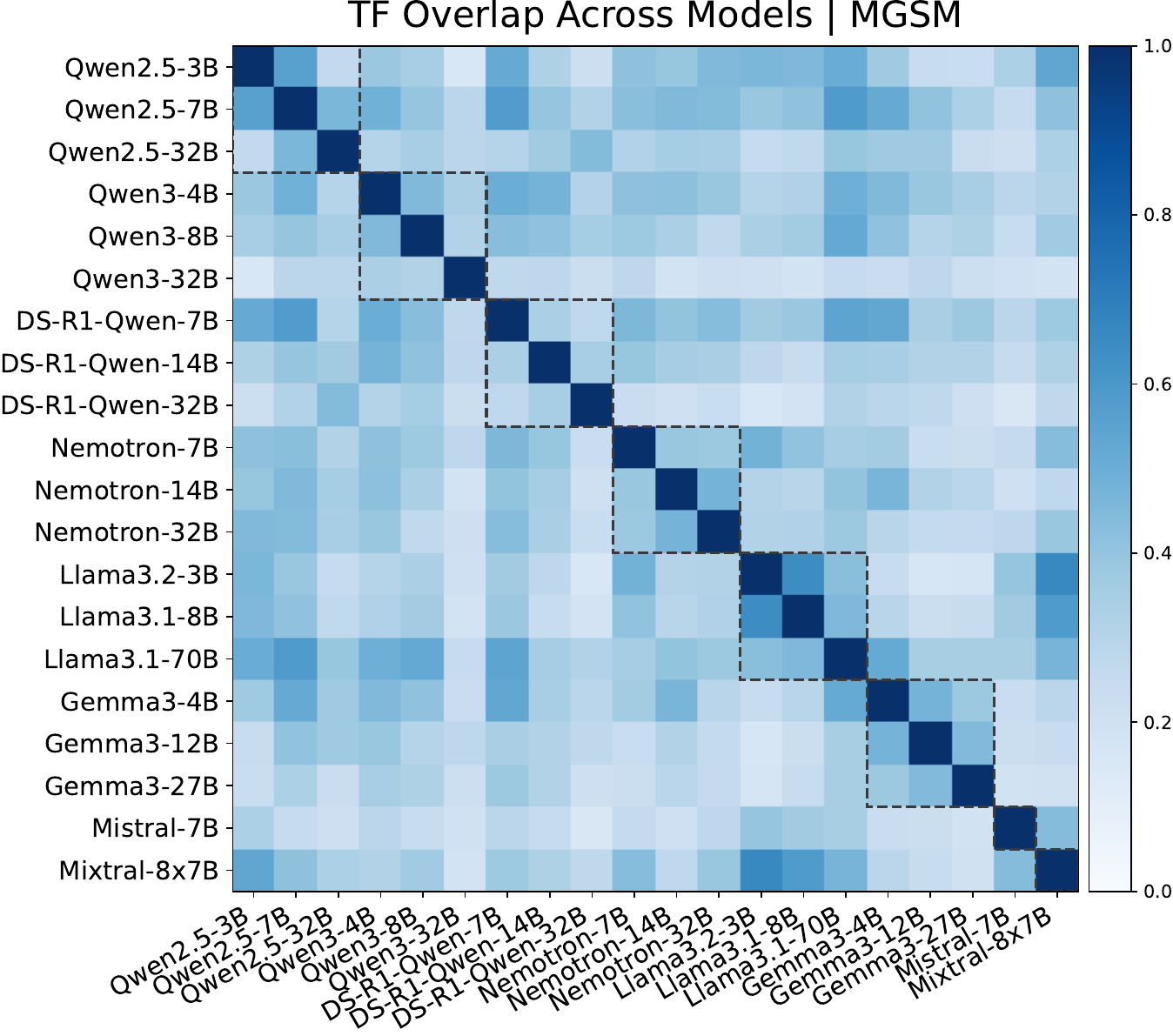}
    \hfill
    \includegraphics[width=0.48\textwidth]{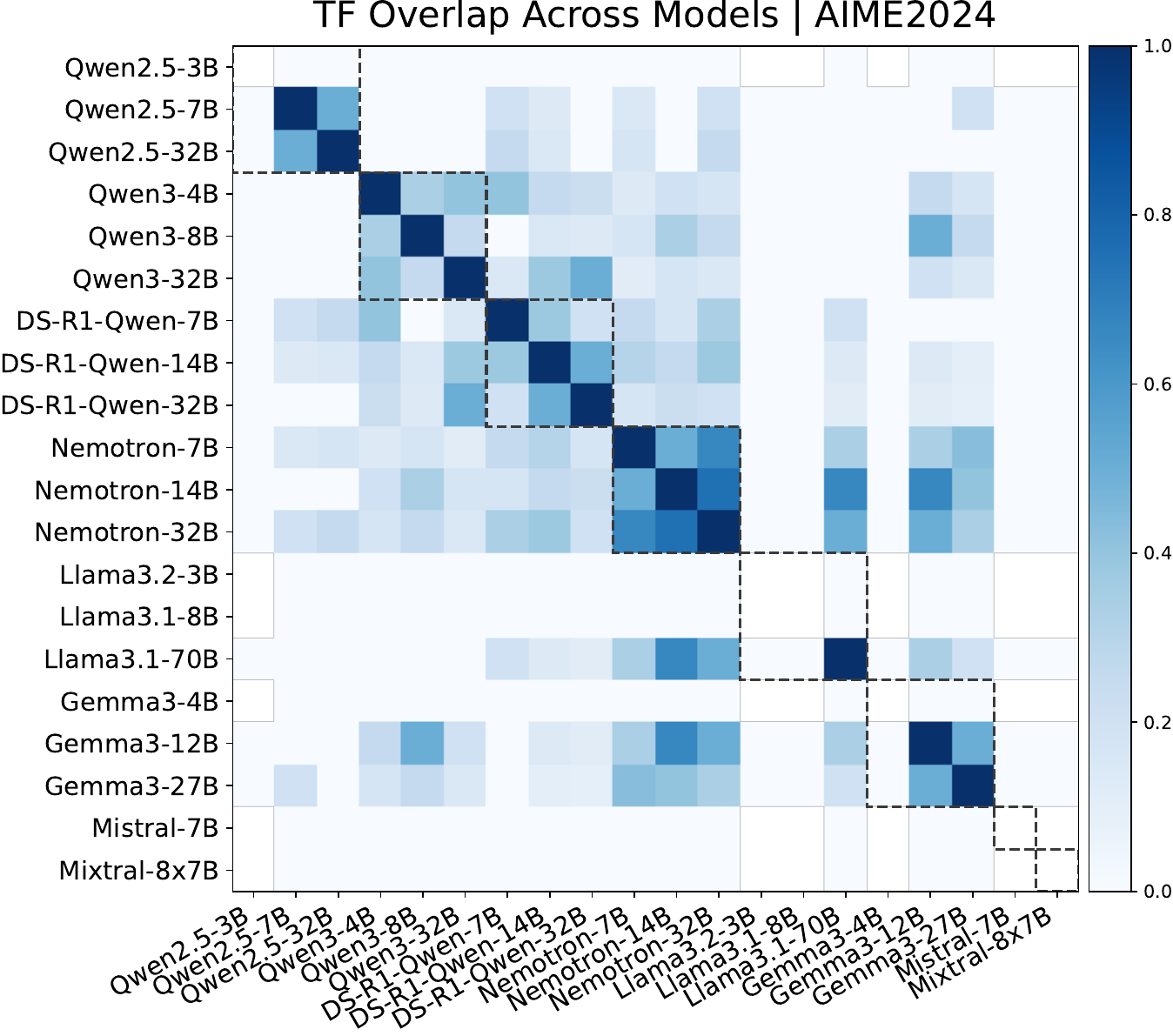}

    \vspace{0.8em}

    \includegraphics[width=0.48\textwidth]{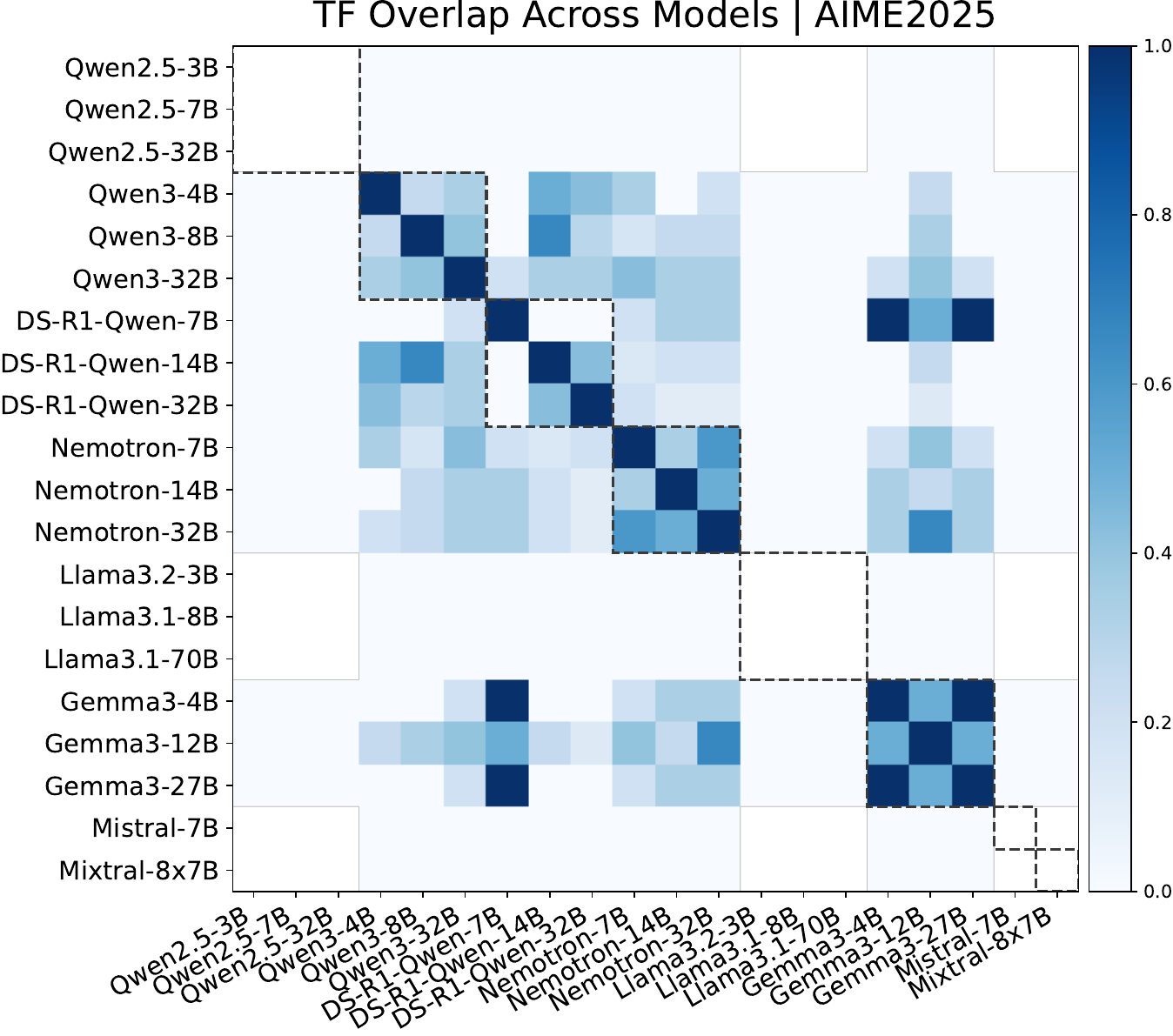}
    \hfill
    \includegraphics[width=0.48\textwidth]{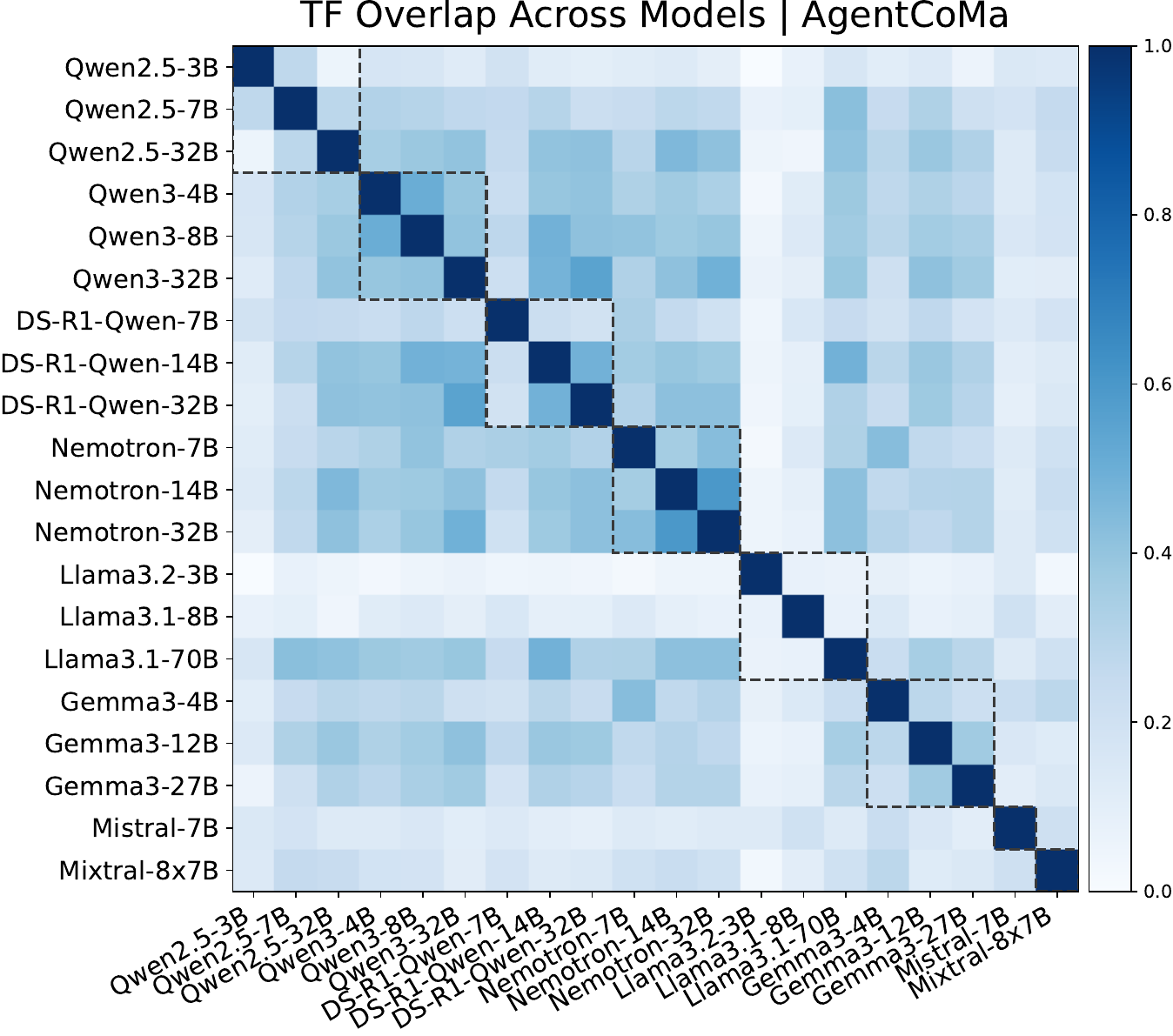}

    \caption{
    Complete TF-overlap heatmaps across all evaluated datasets. 
    From left to right, top to bottom: GSM8K, MATH-500, MGSM, AIME2024, AIME2025, and AgentCoMa.
    Each cell represents the overlap ratio between the TF-transition sets of two models. 
    Darker colors indicate stronger agreement in the reversed problems that induce failures.
    }
    \label{fig:appendix_tf_overlap}
\end{figure*}

Figure~\ref{fig:appendix_tf_overlap} summarizes the TF-overlap results across GSM8K, MATH-500, MGSM, AIME2024, AIME2025, and AgentCoMa.
Each figure shows the pairwise overlap ratio of TF-transition problems between model pairs, where TF denotes cases that are answered correctly on the original problem but incorrectly after applying \framework. 
Higher overlap values indicate that two models tend to fail on similar \problemReversed.

\subsection{Family-level Overlap Significance Analysis}
\label{app:overlap}
We analyze whether \framework-induced failures cluster within model families by measuring TF-index Jaccard similarity between model pairs on correct $\rightarrow$ incorrect transitions. For each dataset, we compare same-family and cross-family model pairs using a one-sided Mann--Whitney U test. For the ``All datasets'' result, we first average TF-index Jaccard scores across datasets for each model pair, and then perform the same test on these pair-level averaged scores to avoid treating repeated observations as independent samples.

As shown in Table~\ref{tab:family_overlap_all}, same-family pairs generally exhibit higher TF-index overlap than cross-family pairs. The aggregated analysis across all datasets remains highly significant, indicating stronger TF-index overlap within model families.
\begin{table}[h]
\centering
\small
\resizebox{\columnwidth}{!}{
\begin{tabular}{lcccc}
\toprule
Dataset & Same-family & Cross-family & $p$-value & Sig. \\
\midrule
AIME2024 & 0.293 & 0.094 & 4.04e-04 & *** \\
AIME2025 & 0.404 & 0.095 & 1.27e-06 & *** \\
AgentCoMa & 0.292 & 0.233 & 7.93e-02 & * \\
GSM8K & 0.144 & 0.104 & 1.49e-01 & \\
MGSM & 0.408 & 0.335 & 2.40e-3 & ***\\
MATH-500 & 0.382 & 0.177 & 4.11e-07 & *** \\
\midrule
\textbf{All datasets} & \textbf{0.284} & \textbf{0.161} & \textbf{4.70e-06} & ***\\
\bottomrule
\end{tabular}
}
\caption{
Family-level overlap analysis of TF transitions induced by \framework. 
We report the mean Jaccard overlap between same-family and cross-family model pairs, together with one-sided Mann--Whitney U test results. 
Statistical significance is denoted as follows: ${}^* p < 0.1$, ${}^{**} p < 0.01$, and ${}^{***} p < 0.001$.
}
\label{tab:family_overlap_all}
\end{table}

\subsection{Failure Decomposition Analysis}
\label{decomposition}
\begin{table*}[h]
\centering
\small
\setlength{\tabcolsep}{5pt}
\begin{tabular}{lccccc}
\toprule
\textbf{Benchmark} &
\textbf{Original Acc.} &
\textbf{Reversed Acc.} &
\textbf{TF Rate} &
\textbf{TF Difficulty} &
\textbf{TF Anchoring Rate} \\
\midrule
AgentCoMa
& 1.6
& 35.2
& 1.3
& 0.93
& 2.1 \\
AIME2024
& 13.9
& 18.1
& 7.3
& 0.79
& 0.7 \\
AIME2025
& 14.0
& 21.9
& 7.8
& 0.80
& 1.1 \\
Math500
& 51.7
& 53.9
& 13.5
& 0.72
& 1.5 \\
MGSM (BN)
& 51.9
& 51.5
& 14.4
& 0.50
& 2.3 \\
MGSM (DE)
& 66.2
& 64.4
& 14.9
& 0.59
& 1.5 \\
MGSM (EN)
& 76.1
& 71.2
& 14.1
& 0.44
& 1.7 \\
MGSM (ES)
& 69.1
& 65.4
& 14.5
& 0.59
& 1.3 \\
MGSM (FR)
& 63.4
& 65.2
& 13.5
& 0.58
& 2.1 \\
MGSM (JA)
& 59.7
& 60.0
& 14.3
& 0.54
& 1.5 \\
MGSM (RU)
& 68.1
& 64.4
& 14.7
& 0.58
& 1.5 \\
MGSM (SW)
& 30.5
& 30.7
& 11.5
& 0.38
& 3.7 \\
MGSM (TE)
& 39.5
& 39.5
& 11.9
& 0.44
& 2.7 \\
MGSM (TH)
& 56.9
& 55.7
& 14.1
& 0.52
& 2.1 \\
MGSM (ZH)
& 65.4
& 62.7
& 15.6
& 0.54
& 1.9 \\
\bottomrule
\end{tabular}
\caption{
Failure decomposition statistics across benchmarks.
All accuracy and rate values are reported as percentages except TF Difficulty, which ranges from 0 to 1.
TF Difficulty measures the average rewritten difficulty of TF cases estimated from cross-model rewritten accuracy.
TF Anchoring Rate measures the proportion of TF failures in which the rewritten prediction exactly matches the original gold answer.
The results suggest that different benchmarks exhibit distinct reversal failure regimes.
}
\label{tab:failure_decomposition_full}
\end{table*}

To better understand the sources of reversal failures (TF), we further analyze whether failures primarily arise from: (1) increased reasoning difficulty in \problemReversed, or (2) attraction toward previously seen answers.

\paragraph{Estimating Rewritten Difficulty.}
For each rewritten instance $x$, we estimate an empirical difficulty score based on cross-model rewritten accuracy:
\[
\mathrm{Difficulty}(x)
=
1 - \mathrm{Acc}_{\mathrm{rew}}(x),
\]
where $\mathrm{Acc}_{\mathrm{rew}}(x)$ denotes the average rewritten accuracy across models for instance $x$. Intuitively, rewritten problems consistently solved by most models receive low difficulty scores, whereas rewritten problems that most models fail to solve receive high difficulty scores.

To reduce circularity when analyzing a particular model family, we additionally compute a leave-one-family-out variant. Specifically, when evaluating failures for a given model family, the difficulty score for each instance is estimated using predictions from all other model families only. This prevents a model family's own errors from artificially inflating the estimated difficulty of the corresponding instances.

We then compare the average rewritten difficulty of TF cases against the overall benchmark-level difficulty distribution. If TF cases are concentrated on high-difficulty rewritten instances, this suggests that reversal failures are at least partially driven by increased reasoning complexity introduced by reversal.

\paragraph{Answer Anchoring Analysis.}
We additionally analyze the extent to which rewritten predictions preserve the original answer. For each rewritten prediction, we define a binary indicator:
\[
\mathrm{SameAsOriginal}
=
\boldsymbol{1}
\left[
\hat{y}_{\mathrm{rew}} = y_{\mathrm{orig}}
\right]
\]
where $\hat{y}_{\mathrm{rew}}$ denotes the rewritten prediction and $y_{\mathrm{orig}}$ denotes the original gold answer.

We further define \emph{anchoring failures} as TF cases in which the rewritten prediction exactly matches the original gold answer:
\[
\mathrm{TF}
\land
(\hat{y}_{\mathrm{rew}} = y_{\mathrm{orig}}).
\]

These cases are particularly informative because the model succeeds on the original problem, fails on the rewritten problem, and nevertheless reproduces the original target answer. We interpret such behavior as evidence of persistent attraction toward previously seen targets or memorized problem-answer patterns.

\paragraph{Interpretation.}
This decomposition allows us to distinguish between at least two qualitatively different sources of reversal failures:

\begin{enumerate}
\item \textbf{Difficulty-driven failures}, where rewritten problems become genuinely harder after reversal;
\item \textbf{Anchoring-driven failures}, where models appear biased toward reproducing original targets despite the reversal objective.
\end{enumerate}

Table~\ref{tab:failure_decomposition_full} presents the failure decomposition across benchmarks.
Empirically, we observe substantially different failure regimes across benchmarks. On AIME-style benchmarks, TF cases exhibit very high rewritten difficulty scores (e.g., $0.79$ on AIME2024 and $0.80$ on AIME2025), while anchoring failure rates remain relatively low ($<1.2\%$). This suggests that reversal failures on these benchmarks are primarily associated with increased reasoning difficulty introduced by reversal.

By contrast, GSM8K-style datasets exhibit noticeably stronger anchoring behavior despite only moderate rewritten difficulty. Across MGSM variants, rewritten difficulty scores typically remain around $0.40$--$0.59$, while TF anchoring failure rates remain consistently nontrivial. For example, the Swahili MGSM variant achieves a TF anchoring failure rate of $3.7\%$ with rewritten difficulty around $0.38$, and the Telugu variant exhibits a TF anchoring failure rate of $2.7\%$ with rewritten difficulty around $0.44$. These gaps suggest that many reversal failures in GSM8K-style settings cannot be explained solely by increased reasoning difficulty, but instead arise from attraction toward original targets.

Math500 appears to lie between these two regimes. Although TF cases exhibit relatively high rewritten difficulty ($0.72$), Math500 still shows nontrivial anchoring behavior ($1.5\%$ TF anchoring failures), suggesting that both increased reasoning complexity and target attraction may contribute simultaneously.

We additionally observe that newer benchmarks such as AIME2025 generally exhibit lower same-as-original overlap than older or synthetic benchmarks. This trend is consistent with the hypothesis that memorized solution trajectories or target-answer associations contribute to reversal failures, since newer benchmarks are less likely to overlap with pretraining corpora.

\section{Details of \framework for RL-Training }
We train both the \textit{Original+Duplicate} and \textit{Original+Reverse} settings using the same hyperparameters shown in Table~\ref{tab:grpo_hyperparams}. 
\label{app:train}
\begin{table}[htbp]
\centering
\small
\setlength{\tabcolsep}{5pt}

\begin{tabular}{lc}
\toprule
Parameter & Value \\
\midrule
Learning Rate & $5 \times 10^{-6}$ \\
Training Epochs & 2 \\
Max Steps & 240 \\
Per-device Batch Size & 1 \\
Gradient Accumulation Steps & 4 \\
Number of Generations & 8 \\
Temperature & 1.2 \\
Max Prompt Length & 2048 \\
Max Completion Length & 16000 \\
LoRA Rank ($r$) & 64 \\
LoRA Alpha & 128 \\
Reward Scaling & Group \\
vLLM Mode & Colocate \\
Use Original Problems & True \\
Use Rewritten Problems & False \\
Original Repeat & 2 \\
Evaluation Ratio & 0.007 \\
\bottomrule
\end{tabular}

\caption{Training hyperparameters for GRPO training.}
\label{tab:grpo_hyperparams}
\end{table}
Specifically, checkpoints are selected based on the best performance on the corresponding evaluation benchmark, since different datasets may favor different stages of training. This protocol avoids bias caused by undertraining or overtraining in either setting and allows us to compare the two training strategies under their respective optimal conditions.
\end{document}